\documentclass[sn-mathphys,Numbered]{sn-jnl}

\usepackage{graphicx}%
\usepackage{multirow}%
\usepackage{amsmath,amssymb,amsfonts}%
\usepackage{amsthm}%
\usepackage{mathrsfs}%
\usepackage[title]{appendix}%
\usepackage{xcolor}%
\usepackage{textcomp}%
\usepackage{manyfoot}%
\usepackage{booktabs}%
\usepackage{algorithm}%
\usepackage{algorithmicx}%
\usepackage{algpseudocode}%
\usepackage{listings}%
\usepackage{rotating} %
\usepackage{makecell}%
\usepackage{natbib}%
\usepackage{setspace}%
\usepackage{hyperref}
\theoremstyle{thmstyleone}%
\theoremstyle{thmstyletwo}%

\theoremstyle{thmstylethree}%

\begin{document}

\title[Article Title]{Forecasting VIX using Bayesian Deep Learning}


\author[1,2]{\fnm{Héctor J.} \sur{Hortúa}}\email{ hhortuao@unbosque.edu.co}

\author[3]{\fnm{Andrés} \sur{Mora}}\email{ a.mora262@uniandes.edu.co}


\affil[1]{\orgdiv{Grupo Signos}, \orgname{Departamento de Matemáticas, Universidad El Bosque},  \city{Bogotá}, \postcode{11001}, \country{Colombia}}

\affil[2]{\orgdiv{Maestría en Estadística Aplicada y Ciencia de Datos}, \orgname{Universidad El Bosque},  \city{Bogotá}, \postcode{11001}, \country{Colombia}}

\affil[3]{\orgdiv{Universidad de los Andes}, \orgname{School of Management}, \orgaddress{\street{Calle 21 \# 1-20}, \city{Bogotá}, \postcode{111711}, \country{Colombia}}}


\abstract{
Recently, deep learning techniques are gradually replacing traditional statistical and machine learning models as the first choice for price forecasting tasks. 
In this paper, we leverage probabilistic deep learning for inferring the volatility index VIX. 
We employ the probabilistic counterpart of WaveNet, Temporal Convolutional Network (TCN), and Transformers. 
We show that TCN outperforms all models with an RMSE around 0.189. 
In addition, it has been well known that modern neural networks provide inaccurate uncertainty estimates. 
For solving this problem, we use the standard deviation scaling to calibrate the networks. 
Furthermore, we found out that MNF with Gaussian prior outperforms Reparameterization Trick and Flipout models in terms of precision and uncertainty predictions.
Finally, we claim that MNF with Cauchy and LogUniform prior distributions yield well calibrated TCN and WaveNet networks being the former that best infer the VIX values.
}




\maketitle

\section{Introduction}\label{sec1}

Investors and regulators are concerned about financial market volatility and crashes. For
this reason, the Volatility index (VIX) was introduced in 1993 by the Chicago Board
Options Exchange (CBOE) with the aim of assessing the expected financial market
volatility in the short-run, i.e. for the next 30 days, since it is calculated as an implied
volatility from the options on the S\&P 500 index on this time-to-maturity \cite{wha}.
The VIX has been proven to be a good predictor of expected stock index shifts, and
therefore as an early warning for investor sentiment and financial market turbulences (see
e.g., \cite{wha}, and more recently, \cite{wang19}). Due to its importance for asset
managers and regulators, it would be useful to foresee the values of the index; however, the
VIX is very difficult to forecast \cite{degia}. There exist several proposals to
predict time series found in the literature classified as conventional and modern methods
(see e.g., \cite{huang21} and the references therein). Among modern methods, deep
learning techniques have been successfully applied to financial time series. Given a
probability space, a time series may be defined as a discrete-time stochastic process, in
other words, a collection of random variables indexed by the integers \cite{qrm}.
Since time series is a sequence of repeated observations of a given set of variables over a
period time \cite{adhi}, where sequences are data points that can be
ordered and past observations may provide relevant information about future ones, deep
learning models employed for other type of sequence models are also useful for time series.
Sequence models may be classified as (see e.g., \cite{kapoor}), (i) one-to-sequence,
where a single input is employed to generate a sequence as an output (e.g., generating text
from an image), (ii) sequence-to-one, where a sequence of data is used to generate a single
output (e.g., sentiment classification), (iii) sequence-to-sequence, a sequential data is the
input to produce a sequence as output (e.g., machine translation). Time series can be
regarded as a special sequence-to-sequence case with trend, seasonality, autocorrelation and
noise characteristics \cite{moroney}. Furthermore, financial time series are characterized
by nonstationary, nonlinear, high-noise, which makes the prediction of these time series
more challenging \cite{huang21}.

Though several deep learning models have been successfully applied to calculate point
estimates of financial variables, all financial models are subject to modeling errors and
uncertainty caused by inexact data inputs, therefore, probabilistic models are more adequate
to achieve more realistic financial inferences and predictions \cite{kanungo}, and then for
optimal decision making \cite{dheur}.  Besides, it has been recently found
that neural networks are miscalibrated \cite{guo17}. Thus, our work intends to tackle
the abovementioned drawbacks by contributing to the literature in the following aspects: (i)
we employ three modern deep learning models to predict the VIX values in a deterministic
framework. These models correspond to WaveNet, Temporal Convolutional Networks
(TCN), and Transformer, (ii) we obtain the probabilistic version of the deterministic models
by using three techniques: Reparameterization Trick (RT), Flipout, and Multiplicative
Normalizing Flows (MNF), (iii) we calibrate the probabilistic models with a
simple approach known as the standard deviation scaling, and finally 
(iv) we find that the probabilistic models of WaveNet-MNF and TCN-MNF with LogUniform and Cauchy priors, respectively, are well calibrated.

The rest of the paper is divided as follows. Section~\ref{sec2} presents an overview of the literature related to the examined models in our study.
Section~\ref{sec3} describe the WaveNet, TCN, and Transformer models.
Section~\ref{sec4} briefly reviews on Bayesian neural networks and the three approaches utilized: Reparameterization Trick (RT), Flipout, and Multiplicative Normalizing Flows (MNF).
Section~\ref{sec5} presents the calibration problem.
Section~\ref{sec6} presents the VIX dataset.
Section~\ref{sec7} explains the methodology of our work. 
Section~\ref{sec8} presents the results of our manuscript on deterministic and probabilistic models and its calibration. 
Finally, Section~\ref{sec9} concludes the paper.

\section{Related Literature}\label{sec2}

Regarding deep learning models applied to financial time series forecasting, \cite{sezer}
performed an exhaustive review of the literature between 2005 and 2019, whereas
\cite{zhang23} carry it out for 2020 and 2022. In these studies, related to VIX,
Psaradellis and Sermpinis \cite{psara} proposed a HAR-GASVR, which is a Heterogeneous
Autoregressive Process (HAR) with Genetic Algorithm with Support Vector Regressor
(GASVR) model. On the other hand, Huang et al.  \cite{huang21} and Yujun et al. \cite{yuju} employ
variational mode decomposition (VMD) methods combined with the long short-term
memory (LSTM) model.

Within the analyzed neural networks in our study, WaveNet has been applied to VIX \cite{boro} and in probabilistic models \cite{sun22}.
In this work, we also implement TCN for financial time series for its adequate performance in time series \cite{bib25}, in financial time series \cite{zhao23}, high-frequency financial data \cite{dai22}, and probabilistic forecasting \cite{bib26}.
Transformer models have been also applied in finance \cite{lopez22} and probabilistic developments for time series \cite{tang21}.

To the best of our knowledge there are few attempts of probabilistic model applications specifically to financial time series \cite{barunik}, \cite{benton}, \cite{du23}.

\section{Neural Networks}\label{sec3}

This section briefly reviews the neural networks employed.
An artificial neural network is a special type of machine learning model that connects neurons organized in layers.
While deep learning model is a kind of neural network with numerous layers and neurons \cite{kapoor}.

\subsection{WaveNet}\label{subsec1}

The WaveNet model was introduced by \cite{oord} in 2016 to generate raw audio waveforms for reproducing human voices and musical instruments purposes. In short, there is a convolutional layer, which access the current and previous inputs. Moreover, there is a stack of dilated (aka atrous) causal one-dimensional convolutional layers, that is, when applying a convolutional layer some input values are omitted, with exponentially increasing filters \cite{gulli}. At the end of the architecture there are dense layers with an adequate activation function. Thus, this model learns short- and long-term patterns. In the original paper, the authors stacked 10 convolutional layers with dilation rates of 1, 2, 4, 8, …, 256, 512 \cite{geron}. Since audio is a type of sequential data, we apply WaveNet to financial time series, which is also a form of sequential data as abovementioned.

\subsection{Temporal Convolutional Network (TCN)}\label{subsec2}

The Temporal Convolutional Network (TCN) was first developed by \cite{bib24} and the authors unified the traditional two-step procedure for video-based action segmentation. The first step involves a Convolutional Neural Network (CNN) that encodes spatial-temporal information, and the second step involves a Recurrent Neural Network (RNN) that captures high-level temporal linkages. Therefore, a TCN may be summarized as a hierarchical temporal encoder-decoder network and allows for long-term patterns, since it is an adaptation of WaveNet \cite{bib24}. The available keras package for TCN coded by Philippe Rémy, and based on \cite{bib27}, is employed in our work.

\subsection{Transformer}\label{subsec3}

The standard Transformer model was developed in \cite{vaswani17}, “Attention is all you need”, which is a non-recurrent encoder decoder architecture that helps to transform (that is why the name Transformer) a sequence into another one. The encoder is generally composed of multi-head attention (MHA) and feed-forward layers with residual connections in between. Though the decoder part is like the encoder, it has a self-attention layer (see e.g., \cite{luong15} for more details about models based on attention). The attention-mechanism is usually represented as Attention(Q, K, V), where Q contains the query, K denotes the keys, and V stands for the values. The main component – MHA – allows for “attending” long-term dependencies in a different way to the short-term dependencies simultaneously. One of its important applications is the Bidirectional Encoder Representations from Transformers (BERT) and GPT-3 models in natural language processing \cite{manu22}. 

\section{Bayesian Neural Networks}\label{sec4}

Probabilistic models like Bayesian Neural Networks (BNN) are more adequate for financial estimates since financial data are prone to measurement errors and are noisy. BNN considers the weights of the network as a probability distribution rather than a single value as in traditional neural networks. To this aim, a prior distribution (in general) over the network weights is placed. Therefore, an appropriate model should quantify the uncertainties to get a better understanding of the risk involved and improve the decision-making process \cite{kanungo}. There are two main uncertainty sources: aleatoric uncertainty (or data uncertainty) and epistemic uncertainty (or model uncertainty) and an ideal BNN would yield more accurate uncertainty estimates because high uncertainties is a sign of imprecise model predictions \cite{bib28}. 
The total uncertainty of a new test output $y^*$ given a new test input $x^*$ may be expressed as (see e.g., \cite{hortua23}, Section 2.2., and the references therein)

\begin{equation}
\widehat{\text{Var}}(y^*|x^*) \approx \frac{1}{T} \sum_{t=1}^{T} \sigma_t^2 + \frac{1}{T} \sum_{t=1}^{T} (\mu_t - \bar{\mu})^2,\label{eqtotunc}
\end{equation}

where $\frac{1}{T} \sum_{t=1}^{T} \sigma_t^2 $, the mean of the prediction variance, represents the aleatoric uncertainty and $\frac{1}{T} \sum_{t=1}^{T} (\mu_t - \bar{\mu})^2$, the variance of the prediction mean, represents  the epistemic uncertainty.

For the inference in probabilistic models, Markov Chain Monte Carlo  (MCMC) approach can be considered (e.g., Metropolis-Hastings, Gibbs sampling, Hamiltonian Monte Carlo – HMC, among others) and variational inference. The latter will be employed in this work and is described as follows (based on \cite{bib29} and its notation, where more details can be found and the references therein).

The output of a BNN is the posterior distribution of the network weights. MCMC methods may be applied to this end; however, they are computationally expensive. Another approach, which is gaining interest in academia is variational inference. Let $p(\omega)$ denote the prior distribution over a parameter $\omega$ (the network weights) on a parameter space $\Omega$. The posterior distribution of the parameter is given by

\begin{equation}
 p(\omega|\mathcal{D}) = \frac{p(\mathcal{D}|\omega)p(\omega)}{p(\mathcal{D})} =\frac{\prod_{i=1}^N p(y_i|x_i,\omega)p(\omega)}{p(\mathcal{D})} ,\label{eq1}
\end{equation}

where,  $p(\mathcal{D}|\omega)$ is known as the likelihood and $p(\mathcal{D})$ the marginal (or evidence) in Bayesian inference framework. In detail, the dataset $\mathcal{D}$ is denoted as $\{(x_i,y_i)\}_{i=1}^N$, where $x_i$ represents the inputs and $y_i$ the outputs of the total $N$ sample of the analyzed dataset.

The goal in variational inference is to find a variational distribution $q_\theta(\omega)$ (indexed by a variational parameter $\theta$ and from a family of distributions $Q$), which approximates to the posterior distribution $p(\omega|\mathcal{D})$. This is done by minimizing the Kullback-Leibler (KL) divergence between the two aforementioned distributions, and it is defined as

\begin{equation}
KL\{q_\theta(\omega)||p(\omega|\mathcal{D})\} := \int_{\Omega}  q_\theta(\omega) \log \frac{q_\theta(\omega)}{p(\omega|\mathcal{D})} d\omega. \label{eq2}
\end{equation}

It can be shown that minimizing the KL divergence is equivalent to maximizing the evidence lower bound (ELBO), which is given by

\begin{equation}
ELBO(q_\theta(\omega) ) = \int_{\Omega}  q_\theta(\omega) \log p(y|x,\omega) d\omega  -  KL\{q_\theta(\omega)||p(\omega)\}. \label{eq3}
\end{equation}

The mean-field approximation with normal distributions may be a proposal for the $Q$ family of distributions \cite{bib30}, \cite{bib31}. That is,

\begin{equation}
q_\theta(\omega) = \prod_{ij}\mathcal{N}(\omega;\mu_{ij},\,\sigma_{ij}^{2}), \label{eq4}
\end{equation}

where $i$ indicates the index of the neurons from the previous layer and $j$ the index of neurons for the current layer. However, it poses a dimensionality problem in the parameters (mean $\mu_{ij}$ and variance $\sigma_{ij}^{2}$) to be estimated. Moreover, the KL divergence may be approximated by sampling the variational distribution, $q_\theta(\omega)$, but it is not possible to perform backpropagation through a random variable. A solution to this problem is Reparameterization Trick, and this is our first approach.

\subsection{Reparameterization Trick}\label{subsec1}

An unbiased and efficient stochastic gradient-based variational inference is provided by (non-local) Reparameterization Trick (RT) and it was applied to variational autoencoders in \cite{bib32} to make backpropagation possible and the output parameters are normally distributed \cite{bib33}, \cite{bib34}. 
Rather than sampling from $\omega$, samples are generated from another variable $\epsilon_{ij}$, which is standard normally distributed, and then $\omega_{ij}=\mu_{ij} + \sigma_{ij}\epsilon_{ij}$ is calculated, allowing for backpropagation.
More details can be found in \cite{bib32}, \cite{bib35},  \cite{bib36} and the TensorFlow documentation at \href{https://www.tensorflow.org/probability/api_docs/python/tfp/layers/DenseReparameterization}{DenseReparameterization}.

\subsection{Flipout}\label{subsec2}

Flipout also provides an unbiased and efficient stochastic gradients estimator, but reduces the variance of the gradient estimates compared to RT. It was proposed by \cite{bib37} and applied to LSTM and convolutional networks. The authors impose two constraints, which are (i) independent perturbations and (ii) these perturbations are centered at zero and it has a symmetric distribution. See more details on the TensorFlow documentation at \href{https://www.tensorflow.org/probability/api_docs/python/tfp/layers/DenseFlipout}{DenseFlipout}

\subsection{Multiplicative Normalizing Flows}\label{subsec3}

Normalizing flows (NF) are probabilistic models useful to fit a complex distribution by learning a transformation (or flow) \cite{bib34}. The NF can be represented as

\begin{equation}
p_{T}(y) = p(x) \left| \det \left( \frac{\partial T(x)}{\partial x} \right) \right|^{-1}, \label{eq:eqNF}
\end{equation}

where $p_{T}(y)$ is the probability density function (pdf) of the transformed variable $y$, \textit{T} is the invertible mapping, and $p(x)$ is the pdf of an invertible random variable (rv) $x$. By including auxiliary rv's $z \sim q_\theta(z)$ and a factorial Gaussian posterior for the weights with mean parameters conditioned on scaling factors that are modelled by NF, the multiplicative normalizing flows (MNF) are obtained \cite{bib38}. Therefore, the variational posterior for fully connected layers (similar result is obtained for convolutional layers) is given by

\begin{equation}
\omega \sim q_z(\omega) = \prod_{ij}\mathcal{N}(\omega;z_i\mu_{ij},\,\sigma_{ij}^{2}), \label{eq6}
\end{equation}

and then a distribution $q(z_K)$ is obtained

\begin{equation}
\log q(z_K) = \log q(z_0) -\sum_{k=1}^K \log \left| \det \left( \frac{\partial f_k}{\partial z_{k-1}} \right) \right|^{-1}, \label{eq7}
\end{equation}

by applying the tranform in Eq. \ref{eq:eqNF} successively as

\begin{equation}
z_K = NF(z_0) = f_K \circ \dots \circ f_1(z_0). \label{eq7}
\end{equation}

Finally, by incorporating an auxiliar distribution $r(z_K|\omega, \phi)$ -- with a new parameter $\phi$ -- the KL divergence may be bounded as follows

\begin{equation}
-KL \left[ q(w) \Vert p(w) \right] \geq \mathbb{E}_{q(w, z_K)} \left[ -KL \left[ q(z_K | w) \Vert p(w) \right] + \log q(z_K) + \log r(z_K | w, \phi) \right].
\label{eq:variational_inequality}
\end{equation}

For more details, see e.g., \cite{hortua23}, Section 2.3. The codes and references found at \href{https://github.com/janosh/tf-mnf}{MNF}are utilized in our work for the MNF model.

\section{Calibration}\label{sec5}

Since the seminal work of \cite{bib39} more attention is being payed in the academia to obtain not only accurate forecasting but also reliable prediction confidence level of robust neural networks. This is achieved by the so-called calibration process.

For classification tasks, it is very well-known calibration techniques such as the Platt calibration, histogram binning, Bayesian binning into quantiles, Temperature scaling, Isotonic regression, ensembled-based calibration methods, and the usual metrics such as expected calibration error (ECE), maximum calibration error (MCE), negative log-likelihood (NLL), and the visual reliability diagrams are employed (see e.g., \cite{bib39}). 
More recently, in the literature, these techniques are classified as post-hoc rescaling of predictions, averaging multiple predictions and data augmentation strategies (\cite{bib40} and the references therein). For a comprehensive revision of calibration methods see \cite{bib41}, \cite{bib42}, \cite{bib43}. 
We follow a similar quantile recalibration method for regression tasks in machine learning \cite{bib44}, and it is seen as a post-hoc rescaling method. 
The standard deviation scaling method (proposed by \cite{bib45}) is adapted in our work, which simply scales the total uncertainty (see Eq. \ref{eqtotunc}) of the uncalibrated network by a factor that minimizes the root mean squared calibration error -- RMSCE -- (\cite{bib46}, Eq. 19).

\section{Data}\label{sec6}

\begin{figure}[h]
\centering
\scalebox{1.0}{\includegraphics[width=\columnwidth]{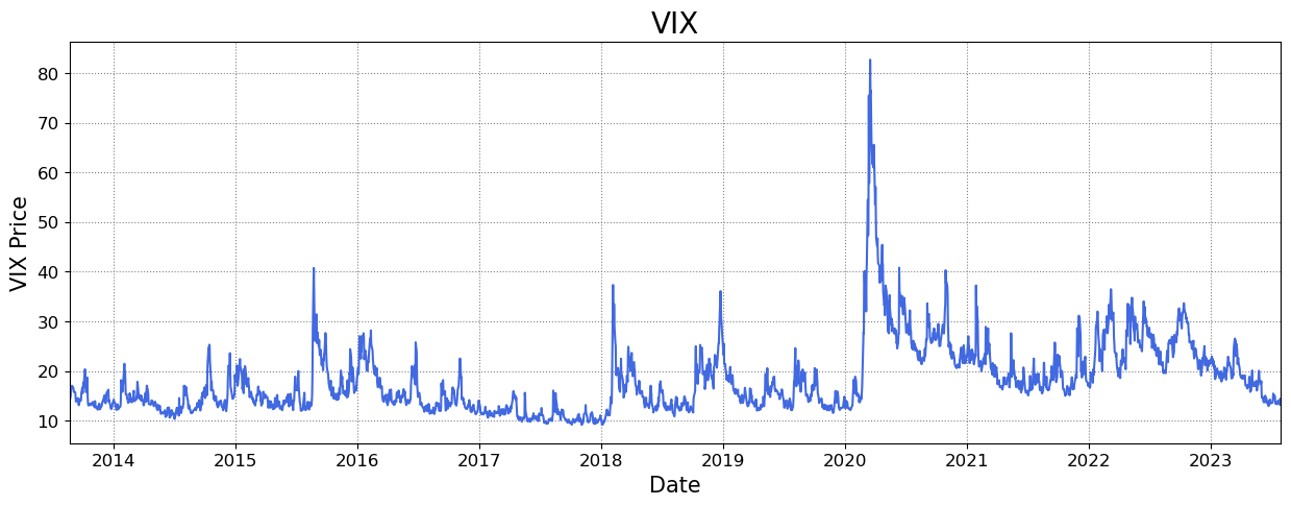}}
\caption{VIX historical price. Daily VIX price taken from August 22, 2013 to July 31, 2023. A peak is observed on March 2020 due to effect of Covid pandemic statement by the World Health Organization (WHO) on financial markets.}
\label{fig:hist_vix}
\end{figure}

Figure~\ref{fig:hist_vix} shows the daily behavior of historical VIX price from August 22, 2013 to July 31, 2023, and its descriptive statistics is presented in Table~\ref{tab:descriptive_statistics}.

\begin{table}[htbp]
\centering
\caption{Descriptive statistics for VIX price. The table shows the usual location and dispersion measures}
\label{tab:descriptive_statistics}
\begin{tabular}{lr}
\toprule
Statistic & Value \\
\midrule
Count       & 2500.00 \\
Mean        & 18.11 \\
Standard deviation         & 7.34 \\
Minimum         & 9.14 \\
25th percentile        & 13.19 \\
50th percentile        & 16.05 \\
75th percentile        & 21.30 \\
Maximum         & 82.69 \\
\bottomrule
\end{tabular}
\end{table}

As seen in the descriptive statistics, the maximum value of the VIX index  was 82.7 on March 2020 as a result of the COVID-19 pandemic. Similar values were recorded in the subprime crisis. The minimum value was 9.14, with a mean of 18.11 and median of 16.05, showing a positive skewness, as seen in the Figure~\ref{fig:histogram_vix}. Values between 15 and 25 are considered moderate, whereas VIX values between 25 and 30 are considered high (have a look at \href{https://www.cboe.com/tradable_products/vix/}{CBOE}),
and this is also confirmed by the boxplot (see Figure~\ref{fig:box_vix}). That is why a robust to outlier scaler transformation of the analyzed data will be employed to train the network models. 
Outliers are observed above the value of 40. As previously mentioned, VIX values greater than 30 are considered extremely high indicating high turbulence in the markets.
Finally the autocorrelation function (ACF) and partial autocorrelation function (PACF) are depicted for the VIX index. See Figure~\ref{fig:acf} and Figure~\ref{fig:pacf}, respectively.
From the serial correlation plot of the VIX time series, a long-term dependence pattern can be observed.
By observing both the ACF and PACF, an AR(2) model could be identified. This is important for traditional time series modelling and for the use of structural time series (STS) modeling in TensorFlow Probability, but this will be the focus of future research.

\section{Methodology}\label{sec7}

The analyzed data consists of the volatility index VIX, downloaded from Yahoo Finance in daily frequency from August 22, 2013 to July 31, 2023. Thus, the total length of data is 2500 observations.
The methodology is described as follows.

In a first step, the VIX time series data is collected from Yahoo Finance, which is freely accessible.  Since time series (with trend, seasonality, autocorrelation, and noise attributes) are a special case of many-to-many sequence domain it is needed a different treatment from the most common tasks in this domain. In particular, the windowed dataset creation as in \cite{moroney} is performed to consider a rolling window for forecasting purposes. We employ a window size of 20 days, i.e. a trading month. Moreover, a robust to outlier scaler transformation of data will be employed. This transformation subtracts the median (instead of the mean as usual) and scales the data to the Interquantile Range (rather than the standard deviation). Furthermore, the split dataset is done in chronological order, 80\% for training set, 10\% for validation set, and 10\% for test set. Thus, we analyze 2000 observations for training, 250 for validation, and 250 for test set, respectively.

Before executing any model, it is important to get a better knowledge of the statistical properties of the analyzed data. Main descriptive statistics (mean, median, standard deviation, first and third quartiles, minimum and maximum) are calculated for the volatility index. In addition, useful graphical tools such as histogram, boxplot, and autocorrelation function (ACF) plots are also obtained.

Then, robust neural network models like WaveNet, TCN, and Transformer will be applied to compare the performance with the usual metrics (MSE, MAE, MSLE, MAPE) for regression tasks and their respective hyperparameters are fine tuned. Bayesian neural networks for each of the deterministic models are obtained by implementing three Bayesian approaches in the last layer of the deterministic model: RT, Flipout and MNF. Finally, the the observed proportion of data falling inside an interval
and the expected proportion of data at different
percentile levels are calculated for each Bayesian neural network and the models are calibrated following the standard deviation scaling. That is, scale the total uncertainty (see Eq. \ref{eqtotunc}) of each model by a factor which minimizes the Root Mean Square Error of Calibration (RMSEC).

The software employed is Python, TensorFlow, Keras Tuner, and TensorFlow Probability. The latter for the probabilistic models. Finally, code repositories for the models and MNF replicability will also be useful in our work.

\section{Results}\label{sec8}

This section presents the results for the deterministic and probabilistic models as its calibration.
We also performed machine learning techniques to forecast the VIX price and the results are found in Table~\ref{table:pycaret}. 

\begin{sidewaystable}[h!]
\caption{Results of Machine Learning techniques to forecast the VIX price obtained by utilizing PyCaret. The table shows the traditional metrics for regression tasks}
\label{table:pycaret}
\centering
\begin{tabular}{lcccccccc}
\hline
Model & MASE & RMSSE & MAE & RMSE & MAPE & SMAPE & R2 & TT (Sec) \\ 
\hline
Naive Forecaster & 0.0548 & 0.0371 & 0.2111 & 0.2402 & 0.0154 & 0.0154 & -0.9995 & 2.2767 \\
ETS & 0.0840 & 0.0524 & 0.3237 & 0.3395 & 0.0238 & 0.0237 & -7.4799 & 1.2267 \\
Auto ARIMA & 0.0864 & 0.0561 & 0.3329 & 0.3633 & 0.0245 & 0.0243 & -9.0459 & 71.5200 \\
Exponential Smoothing & 0.0902 & 0.0582 & 0.3474 & 0.3775 & 0.0255 & 0.0255 & -7.9595 & 0.3000 \\
Theta Forecaster & 0.0987 & 0.0641 & 0.3801 & 0.4153 & 0.0279 & 0.0278 & -11.8816 & 0.0833 \\
Huber w/ Cond. Deseasonalize \& Detrending & 0.1415 & 0.0876 & 0.5448 & 0.5679 & 0.0401 & 0.0391 & -30.8392 & 0.1133 \\
Croston & 0.1845 & 0.1144 & 0.7107 & 0.7416 & 0.0525 & 0.0508 & -39.9722 & 0.0533 \\
Linear w/ Cond. Deseasonalize \& Detrending & 0.2013 & 0.1279 & 0.7752 & 0.8293 & 0.0571 & 0.0546 & -73.2645 & 0.9400 \\
Ridge w/ Cond. Deseasonalize \& Detrending & 0.2013 & 0.1279 & 0.7752 & 0.8293 & 0.0571 & 0.0546 & -73.2679 & 0.7100 \\
Bayesian Ridge w/ Cond. Deseasonalize \& Detrending & 0.2034 & 0.1291 & 0.7835 & 0.8365 & 0.0577 & 0.0551 & -73.9824 & 0.1467 \\
Orthogonal Matching Pursuit w/ Cond. Deseasonalize \& Detrending & 0.2137 & 0.1343 & 0.8229 & 0.8704 & 0.0605 & 0.0582 & -62.1286 & 0.1100 \\
Seasonal Naive Forecaster & 0.2475 & 0.1924 & 0.9533 & 1.2471 & 0.0702 & 0.0646 & -207.5962 & 1.9700 \\
Elastic Net w/ Cond. Deseasonalize \& Detrending & 0.2511 & 0.1554 & 0.9671 & 1.0074 & 0.0711 & 0.0681 & -83.0503 & 0.5200 \\
Extreme Gradient Boosting w/ Cond. Deseasonalize \& Detrending & 0.2801 & 0.1890 & 1.0790 & 1.2249 & 0.0795 & 0.0750 & -104.2627 & 1.1367 \\
Light Gradient Boosting w/ Cond. Deseasonalize \& Detrending & 0.2902 & 0.1856 & 1.1173 & 1.2026 & 0.0818 & 0.0780 & -56.9334 & 0.5233 \\
Lasso w/ Cond. Deseasonalize \& Detrending & 0.2927 & 0.1804 & 1.1270 & 1.1693 & 0.0828 & 0.0789 & -103.0531 & 0.2467 \\
Random Forest w/ Cond. Deseasonalize \& Detrending & 0.2985 & 0.1894 & 1.1498 & 1.2278 & 0.0849 & 0.0792 & -173.5435 & 4.9500 \\
STLF & 0.3110 & 0.2190 & 1.1978 & 1.4193 & 0.0881 & 0.0825 & -132.4984 & 0.1133 \\
Gradient Boosting w/ Cond. Deseasonalize \& Detrending & 0.3140 & 0.1999 & 1.2092 & 1.2959 & 0.0889 & 0.0843 & -113.4597 & 0.8800 \\
Extra Trees w/ Cond. Deseasonalize \& Detrending & 0.3387 & 0.2120 & 1.3044 & 1.3740 & 0.0959 & 0.0904 & -128.1481 & 3.2167 \\
CatBoost Regressor w/ Cond. Deseasonalize \& Detrending & 0.3863 & 0.2473 & 1.4876 & 1.6025 & 0.1089 & 0.1024 & -107.7005 & 4.2967 \\
Decision Tree w/ Cond. Deseasonalize \& Detrending & 0.4900 & 0.3689 & 1.8857 & 2.3901 & 0.1366 & 0.1117 & -691.7551 & 3.5000 \\
ARIMA & 0.5536 & 0.3767 & 2.1318 & 2.4417 & 0.1561 & 0.1658 & -255.7048 & 0.4367 \\
Grand Means Forecaster & 1.1619 & 0.6910 & 4.4739 & 4.4782 & 0.3282 & 0.2817 & -1011.7255 & 2.7733 \\
AdaBoost w/ Cond. Deseasonalize \& Detrending & 1.2890 & 0.7661 & 4.9635 & 4.9651 & 0.3639 & 0.3074 & -1152.4047 & 0.2400 \\
Polynomial Trend Forecaster & 2.5575 & 1.5198 & 9.8477 & 9.8496 & 0.7220 & 0.5303 & -4851.9805 & 0.0367 \\
Lasso Least Angular Regressor w/ Cond. Deseasonalize \& Detrending & 2.5705 & 1.5277 & 9.8977 & 9.9011 & 0.7258 & 0.5322 & -4942.8657 & 0.1733 \\
\hline
\end{tabular}
\end{sidewaystable}

Interestingly, the Naive Forecaster approach, which basically assumes that future values will behave similarly as past values, is the best model followed by the Exponential Smoothing (ETS) algorithm.
In particular, we follow the PyCaret tutorial for time series found at \href{https://github.com/pycaret/pycaret/blob/master/tutorials/Tutorial\%20-\%20Time\%20Series\%20Forecasting.ipynb}{Pycaret-Github} and more details are found at \href{https://pycaret.gitbook.io/docs}{Pycaret-Doc}.

\subsection{Deterministic Models}\label{subsec1}

After tuning the hyperparameters for the WaveNet model, the following values are obtained: seven (7) blocks, five (5) layers per block, and 96 filters. For more specific details about the code see \href{https://github.com/ageron/handson-ml2/}{geron-github} and \href{https://github.com/sinusgamma/probabilistic_wavenet_fx/blob/master/wavenet_fx_final.ipynb}{wavenet}.

While for the TCN model, we found one stack (nb\_stack),  and 64 filters to use in the convolutional layer (nb\_filters). The same number of units (64) is fixed for the LSTM, which is the layer that connects after the TCN architecture, the setup of [1, 2, 4, 8, 16] for the dilations (dilation\_list), and the kernel size is equal to 3. See more details at \href{https://github.com/philipperemy/keras-tcn}{tcn}.
 
For the Transformer model, the Keras documentation for time series classification is adapted in our work. 
In the MHA part, we found 256 units for the size of each attention head for query and key (key\_dim), eight (8) attention heads (num\_heads), and dropout probability of 0.10, according to the Grid Search run in Keras Tuner.
While, in the feed forward part, the number of filters (ff\_dim) of eight(8) are utilized in the one dimensional convolutional layer.
Moreover, we stack eight (8) of these transformer enconder blocks.
Finally, for the multilayer perceptron head, 264 units and a dropout probability of 0.10 are employed.
For more details, have a look at the Keras documention: \href{https://keras.io/api/layers/attention_layers/multi_head_attention/}{MHA} and \href{https://keras.io/examples/timeseries/timeseries_classification_transformer/}{transformer}.

\begin{figure}[h]
\centering
\scalebox{0.7}{\includegraphics[width=\columnwidth]{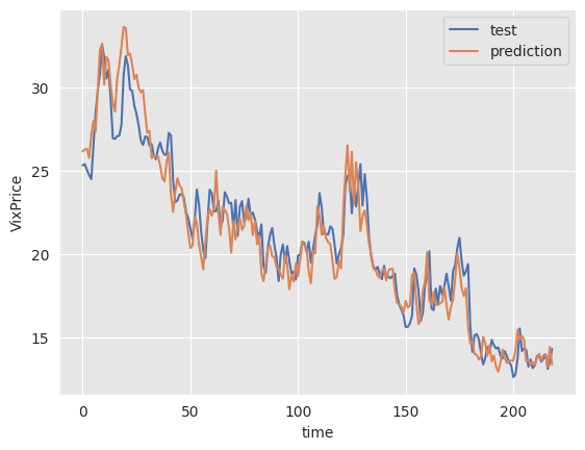}}
\caption{Prediction of the deterministic WaveNet model for VIX test dataset. A good fit of the model is observed except for the peaks at the beginning of the graph.}
\label{fig:wn_det}
\end{figure}

\begin{figure}[h]
\centering
\scalebox{0.7}{\includegraphics[width=\columnwidth]{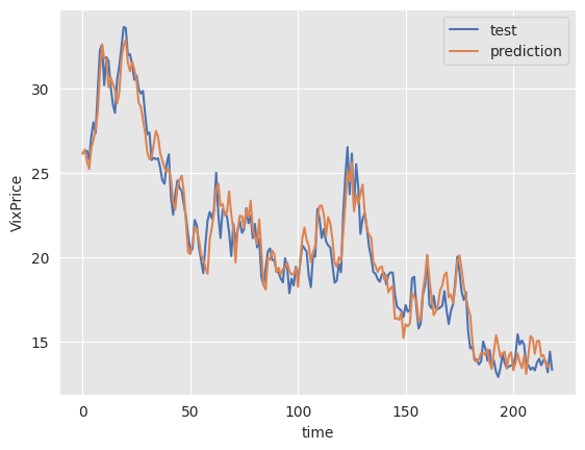}}
\caption{Prediction of the deterministic TCN model for VIX test dataset. A good fit of the model is observed except for the low values of the VIX at the end of the graph.}
\label{fig:tcn_det}
\end{figure}

\begin{figure}[h]
\centering
\scalebox{0.7}{\includegraphics[width=\columnwidth]{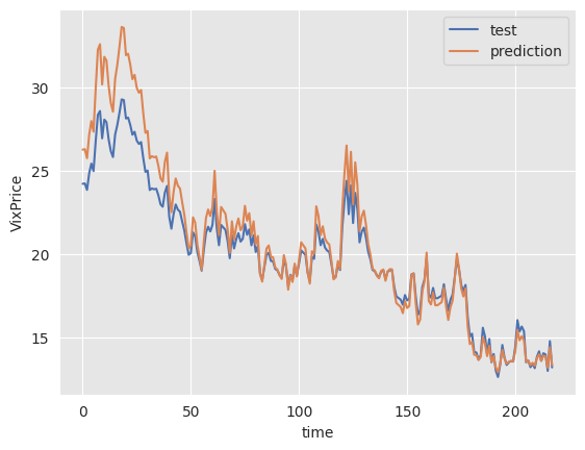}}
\caption{Prediction of the deterministic Transformer model for VIX test dataset. A good fit of the model is observed except for the peaks at the beginning of the graph.}
\label{fig:transf_det}
\end{figure}

The Table~\ref{table:performance_metrics} exhibits the metrics for training, validation, and test set for the three models: Wavenet, TCN, and Transformer. 
The Transformer model is the network with the miminum loss (Huber Loss) in test set, while the TCN presents the lower values for MAE, RMSE, and MSLE, and the WaveNet exhibits the minimum MAPE.
Furthermore, Figure~\ref{fig:wn_det}, Figure~\ref{fig:tcn_det} and Figure~\ref{fig:transf_det} present the results of the prediction and actual data for the WaveNet, TCN, and Transformer models, respectively.
In a visual analysis, the TCN seems to be the model that fits the best to the test dataset.
For lower VIX values, i.e. in the las part of the plot, the TCN does not predict adequately the actual data, but the WaveNet and Transformer do a good job.
However, the WaveNet behaves better than the Transformer for higher values of VIX, that is, at the very beginning of the graph.

\begin{sidewaystable}[h!]
\centering
\caption{Performance metrics for deterministic models. Usual metrics for regression tasks are calculated for train set, valid set, and test set}
\label{table:performance_metrics}
\begin{tabular}{|l|ccccc|ccccc|ccccc|}
\hline
\multicolumn{1}{|c|}{\textbf{Models}} & \multicolumn{5}{c|}{\textbf{Train set}} & \multicolumn{5}{c|}{\textbf{Validation set}} & \multicolumn{5}{c|}{\textbf{Test set}} \\ \hline
 & Loss & MAE & RMSE & MAPE & MSLE & Loss & MAE & RMSE & MAPE & MSLE & Loss & MAE & RMSE & MAPE & MSLE \\ \hline
Wavenet & 0.043 & 0.192 & 0.312 & 162.555 & 0.017 & 0.061 & 0.263 & 0.352 & 29.982 & 0.022 & 0.020 & 0.159 & 0.200 & 48.497 & 0.010 \\
TCN & 0.053 & 0.174 & 0.437 & 117.531 & 0.012 & 0.071 & 0.279 & 0.382 & 28.781 & 0.025 & 0.018 & 0.145 & 0.189 & 110.247 & 0.008 \\ 
Transformer & 0.159 & 0.398 & 0.737 & 246.745 & 0.071 & 0.241 & 0.553 & 0.723 & 75.507 & 0.093 & 0.010 & 0.337 & 0.449 & 159.991 & 0.039 \\ \hline
\end{tabular}
\end{sidewaystable}

\subsection{Probabilistic Models}\label{subsec2}

The Bayesian techniques of RT, Flipout and MNF reviewed in Section~\ref{sec4} are employed in the last layer of the previous deterministic networks to obtain their respective probabilistic models.

The Table~\ref{table:probabilistic_performance_metrics}  presents the metrics for the probabilistic models and for sake of comparison only the test dataset results will be considered in our analysis.
\begin{itemize}
\item For the WaveNet case, the MNF is the model with the lowest value of loss and RMSE, and similar MAE and MSLE values are obtained for MNF and Flipout, and Flipout performs the best for MAPE.
RT is outperformed in most of the metrics by the other two models.
\item MNF has the minimum MAE, RMSE, and MSLE values for the TCN network, whereas RT outperforms in loss and MAPE metrics, and Flipout performs the worst in most of the metrics.
\item The results of the metrics for the Transformer network show that MNF has the lowest MAPE value, RT for MAE, RMSE, and MSLE, and Flipout for the loss metric.
\end{itemize}
As a consequence, despite the mixed results in the different models, it is observed a good performance of the MNF model in general. An important result of \cite{guo17} is that neural networks are miscalibrated and this affects the forecasting performance of a model. The next section deals with this issue, the calibration problem.

\begin{sidewaystable}[h!]
\centering
\caption{Performance metrics for probabilistic models. Usual metrics for regression tasks are calculated for train set, valid set, and test set}
\label{table:probabilistic_performance_metrics}
\begin{tabular}{|l|ccccc|ccccc|ccccc|}
\hline
\multicolumn{1}{|c|}{\textbf{Models}} & \multicolumn{5}{c|}{\textbf{Train set}} & \multicolumn{5}{c|}{\textbf{Validation set}} & \multicolumn{5}{c|}{\textbf{Test set}} \\ \hline
 & Loss & MAE & RMSE & MAPE & MSLE & Loss & MAE & RMSE & MAPE & MSLE & Loss & MAE & RMSE & MAPE & MSLE \\ \hline
\multicolumn{16}{|c|}{\textbf{Wavenet}} \\ \hline
RT & 0.250 & 0.312 & 0.406 & 258.558 & 0.029 & 0.831 & 0.481 & 0.634 & 65.607 & 0.078 & 0.556 & 0.394 & 0.513 & 407.712 & 0.062 \\
Flipout & 0.286 & 0.325 & 0.462 & 268.850 & 0.029 & 0.699 & 0.413 & 0.519 & 67.202 & 0.053 & 0.315 & 0.333 & 0.419 & 211.159 & 0.041 \\
MNF & 0.158 & 0.324 & 0.589 & 262.369 & 0.031 & 0.551 & 0.376 & 0.490 & 48.623 & 0.047 & 0.212 & 0.333 & 0.415 & 409.240 & 0.042 \\ \hline
\multicolumn{16}{|c|}{\textbf{TCN}} \\ \hline
RT & 0.495 & 0.324 & 0.761 & 170.072 & 0.053 & 1.034 & 0.461 & 0.641 & 42.238 & 0.075 & 0.540 & 0.326 & 0.458 & 156.208 & 0.037 \\
Flipout & 0.689 & 0.402 & 0.765 & 316.353 & 0.054 & 0.989 & 0.440 & 0.565 & 54.654 & 0.064 & 0.635 & 0.349 & 0.423 & 485.183 & 0.043 \\
MNF & 0.632 & 0.359 & 0.698 & 220.614 & 0.044 & 0.946 & 0.392 & 0.505 & 61.568 & 0.053 & 0.604 & 0.307 & 0.384 & 166.708 & 0.033 \\ \hline
\multicolumn{16}{|c|}{\textbf{Transformers}} \\ \hline
RT          & 1.991 & 0.578 & 0.887 & 340.180 & 0.115 & 2.364 & 0.745 & 0.976 & 89.016 & 0.161 & 1.931 & 0.604 & 0.782 & 378.655 & 0.136 \\
Flipout     & 1.362 & 0.579 & 0.909 & 322.152 & 0.112 & 1.728 & 0.766 & 1.012 & 102.733 & 0.179 & 1.302 & 0.664 & 0.889 & 355.455 & 0.144 \\
MNF   & 3.147 & 0.592 & 0.916 & 408.669 & 0.116 & 3.463 & 0.811 & 1.025 & 124.970 & 0.183 & 3.111 & 0.645 & 0.827 & 153.303 & 0.140 \\ \hline
\end{tabular}
\end{sidewaystable}

\subsection{Calibration}\label{subsec3}
This work implements three robust neural networks (WaveNet, TCN, and Transformer) mostly employed in the literature for many-to-many sequence tasks.
After having the hyperparameters fine-tuned, these networks have been trained for the VIX forecasting purposes with good results in a deterministic manner.
As mentioned in the Introduction Section, probabilistic models are more appropriate to achieve more realistic financial inferences and predictions.
To this aim, we implement three models: RT, Flipout, and MNF in the last layer of the deterministic models and calculate their respective (total) uncertainties (see Eq. \ref{eqtotunc}).
However, these models are miscalibrated and affect not only the point estimates but also the uncertainty around these point predictions.

To analyze (mis)calibration, the observed proportion of data falling inside an interval and the expected proportion of data of a standard normal distribution at different percentile levels (i.e., 10\%, 20\%, 30\%, 40\%, 50\%, 60\%, 70\%, 80\%, and 90\%) are calculated.
Then, we plot the observed proportion of data vs the expected proportion of data (as per in \cite{bib46}, Fig. 12-b), before and after the calibration.
This graph resembles a modified reliability diagram for classification tasks.
A miscalibration is evidenced in the aforementioned plot, if the observed proportion of data lie far from the diagonal of the graph.
On the other hand, a perfect calibration is noticed when all the observed proportion of data lies in the diagonal.

If a network model is miscalibrated, a post-hoc rescaling method is followed to calibrate the model.
In other words, the total uncertainty (see Eq. \ref{eqtotunc}) of the miscalibrated model is multiplied by a factor $c$ that minimizes the RMSCE \cite{bib46}, Eq. 19, given by

\begin{equation}
RMSCE = \sqrt{\mathbb{E}_{p \in [0,1]} \left( p - c*\hat{p}(p) \right)^2}, 
\label{eq:rmsce}
\end{equation}

where $p$ is the expected proportion of data and $\hat{p}(p)$ is the observed proportion of data that lies inside the calculated interval given by the total uncertainty.

\begin{figure}[h!]
\centering
\scalebox{0.8}{\includegraphics[width=\columnwidth]{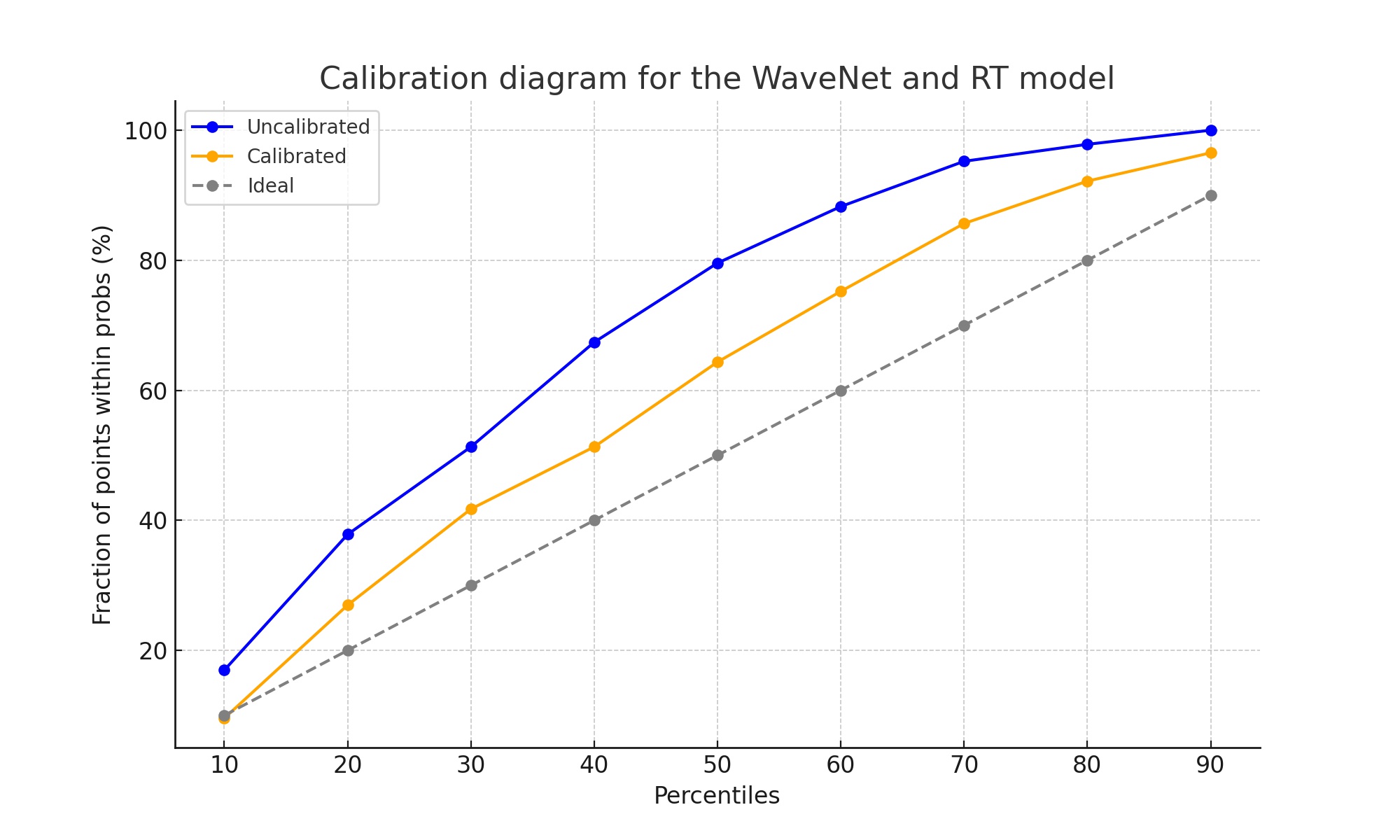}}
\caption{Calibration diagram for the WaveNet with RT model. After minimizing the RMSCE, the scaling factor is equal to 0.7373. The dashed diagonal line represents a perfect calibration.}
\label{fig:wn_rt_calib}
\end{figure}

\begin{figure}[h!]
\centering
\scalebox{0.7}{\includegraphics[width=\columnwidth]{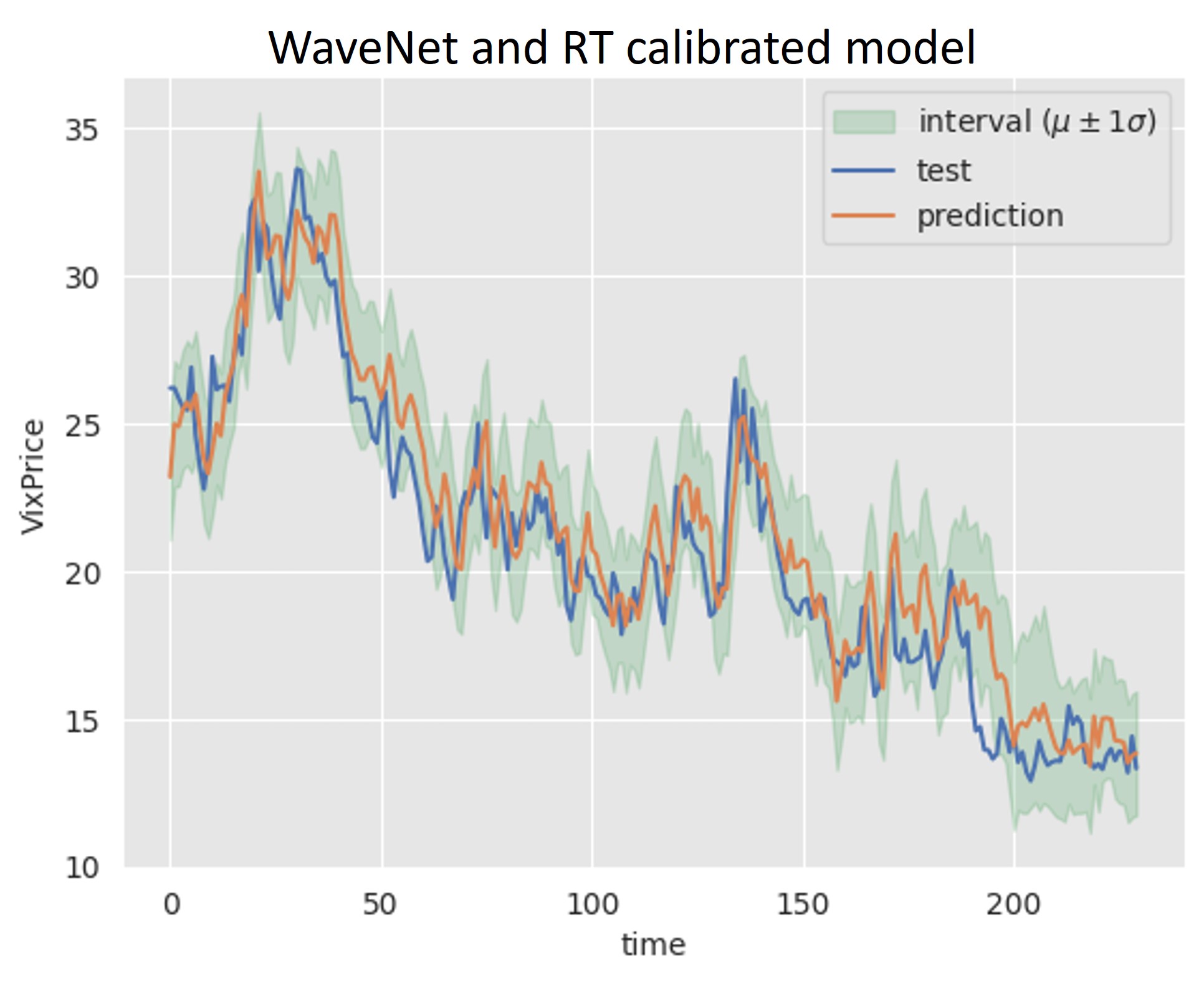}}
\caption{Prediction of the probabilistic WaveNet and RT model for VIX test dataset}\label{fig:wn_rt_prob}
\end{figure}

\begin{figure}[h!]
\centering
\scalebox{0.8}{\includegraphics[width=\columnwidth]{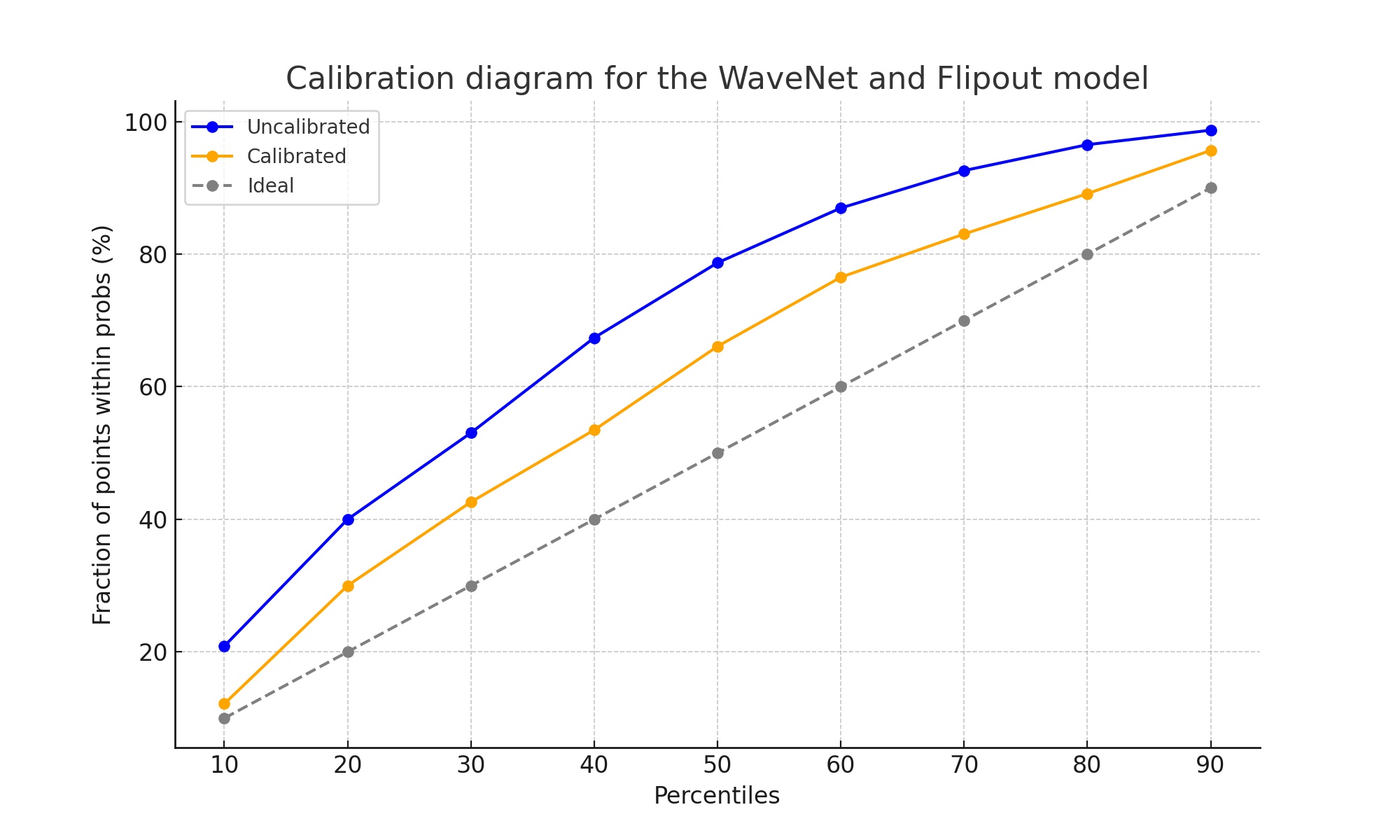}}
\caption{Calibration diagram for the WaveNet with Flipout model. After minimizing the RMSCE, the scaling factor is equal to 0.7392. The dashed diagonal line represents a perfect calibration.}
\label{fig:wn_ft_calib}
\end{figure}

\begin{figure}[h!]
\centering
\scalebox{0.7}{\includegraphics[width=\columnwidth]{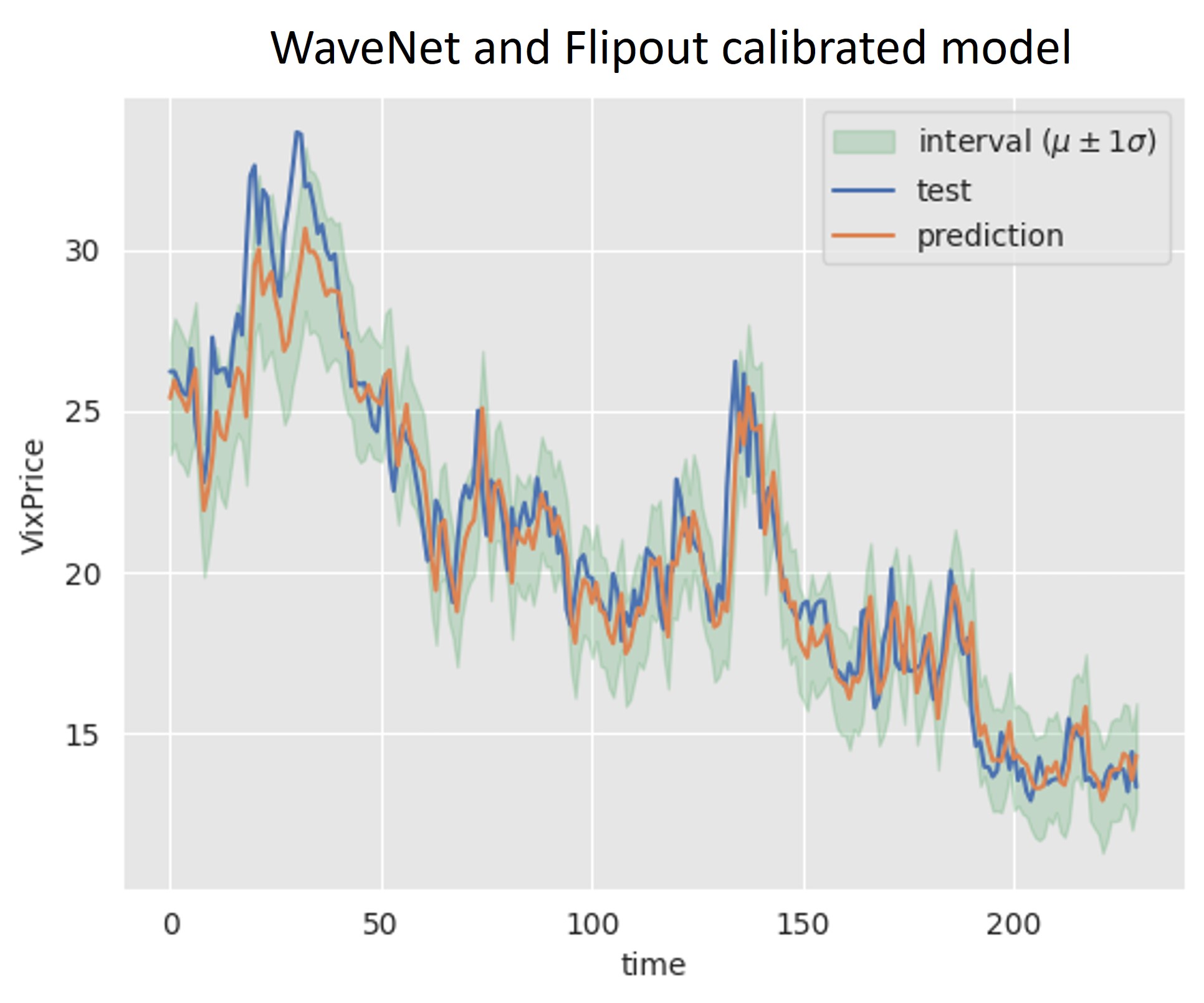}}
\caption{Prediction of the probabilistic WaveNet and Flipout model for VIX test dataset}\label{fig:wn_ft_prob}
\end{figure}

\begin{figure}[h!]
\centering
\scalebox{0.8}{\includegraphics[width=\columnwidth]{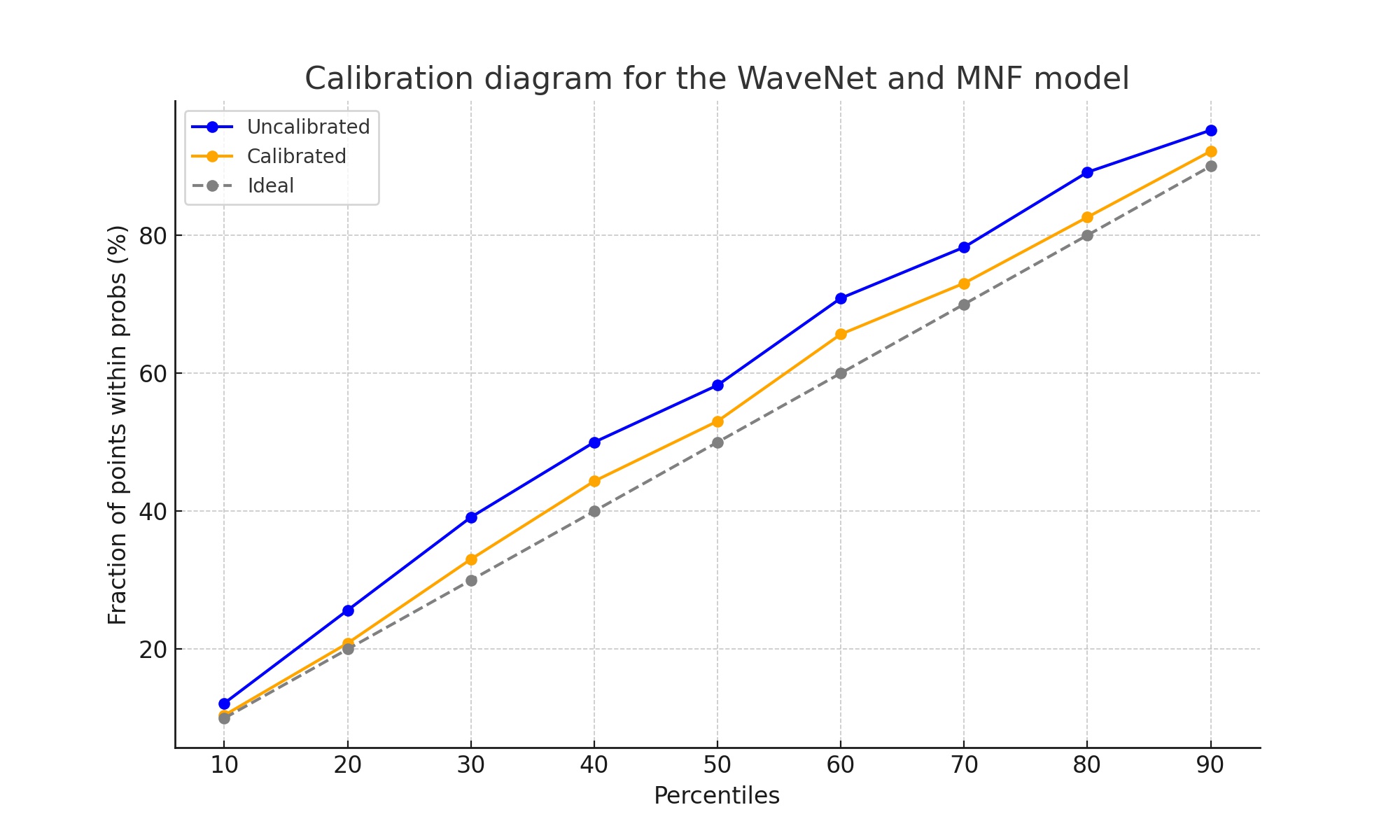}}
\caption{Calibration diagram for the WaveNet with MNF model. After minimizing the RMSCE, the scaling factor is equal to 0.8836. The dashed diagonal line represents a perfect calibration.}
\label{fig:wn_mnf_calib}
\end{figure}

\begin{figure}[h!]
\centering
\scalebox{0.7}{\includegraphics[width=\columnwidth]{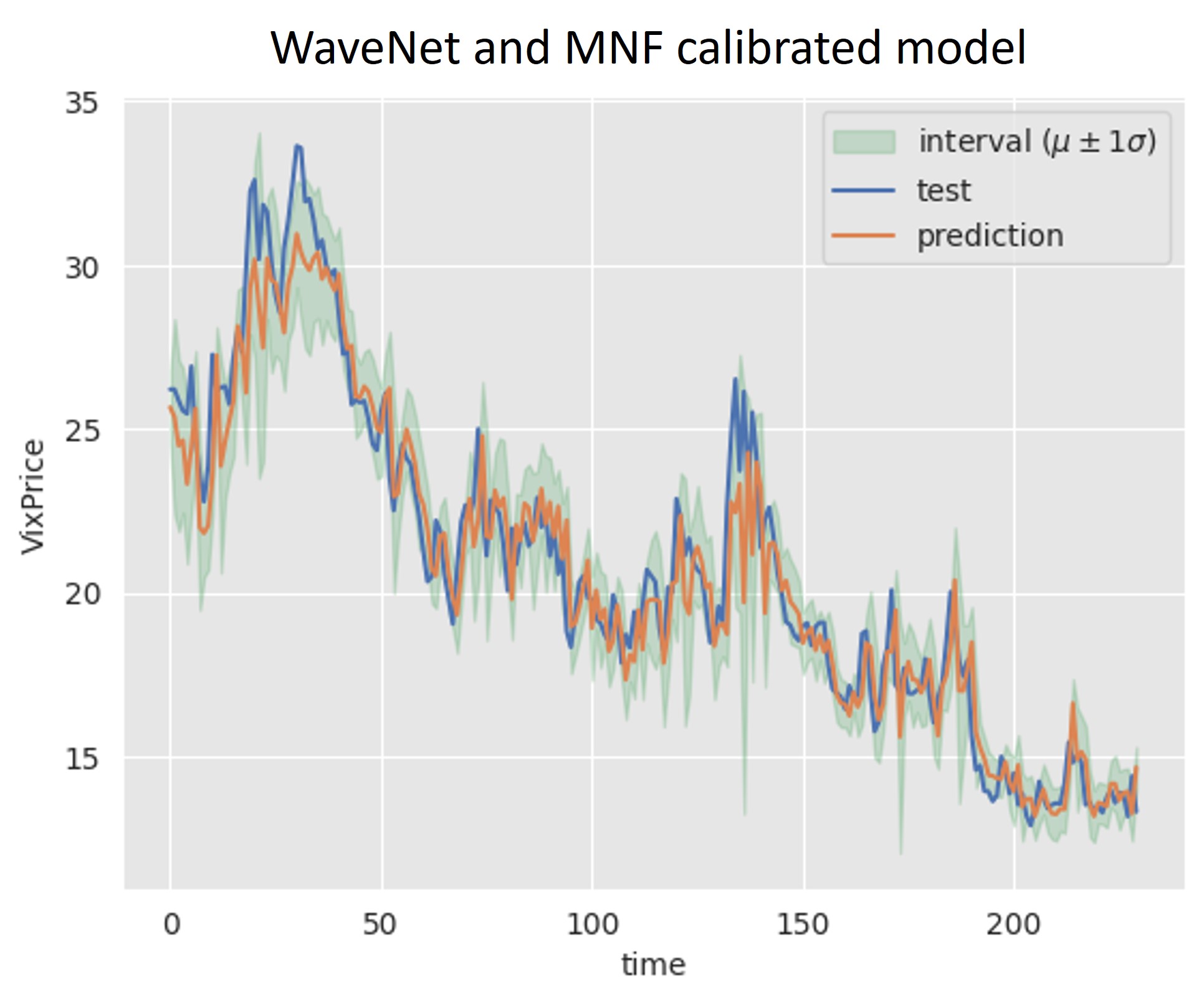}}
\caption{Prediction of the probabilistic WaveNet and MNF model for VIX test dataset}\label{fig:wn_mnf_prob}
\end{figure}

\begin{figure}[h!]
\centering
\scalebox{0.8}{\includegraphics[width=\columnwidth]{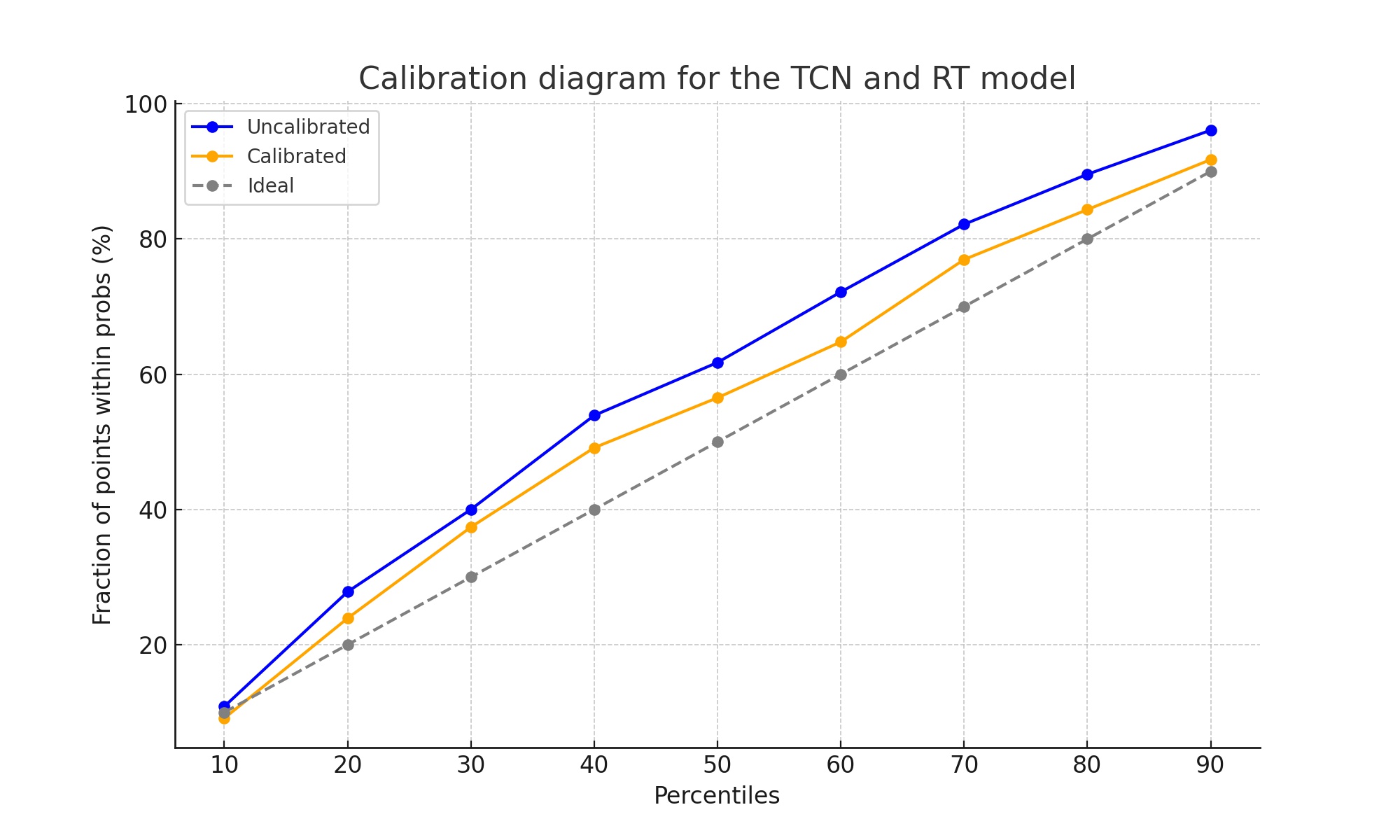}}
\caption{Calibration diagram for the TCN with RT model. After minimizing the RMSCE, the scaling factor is equal to 0.8589. The dashed diagonal line represents a perfect calibration.}
\label{fig:tcn_rt_calib}
\end{figure}

\begin{figure}[h!]
\centering
\scalebox{0.7}{\includegraphics[width=\columnwidth]{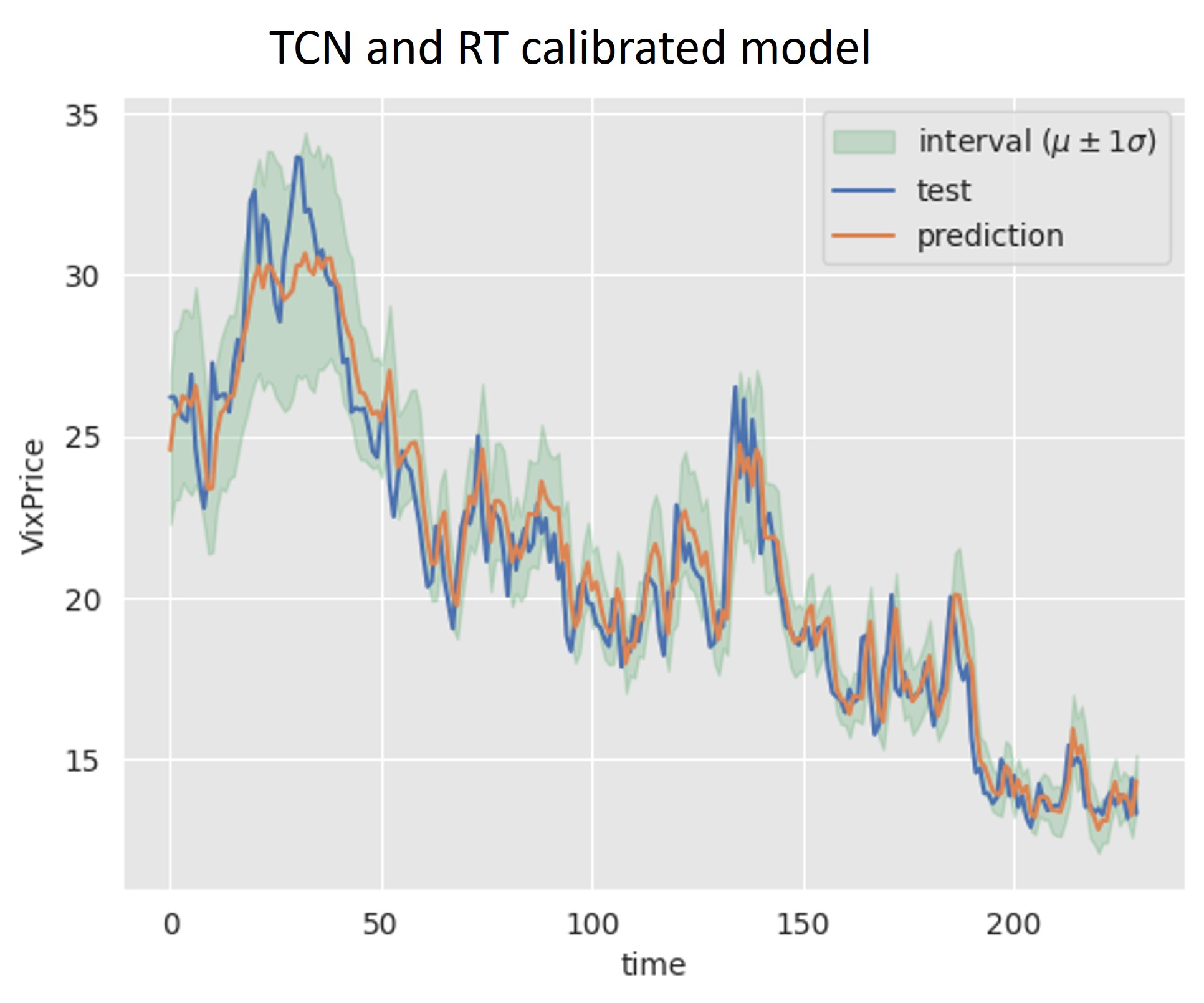}}
\caption{Prediction of the probabilistic TCN and RT model for VIX test dataset}\label{fig:tcn_rt_prob}
\end{figure}

\begin{figure}[h!]
\centering
\scalebox{0.8}{\includegraphics[width=\columnwidth]{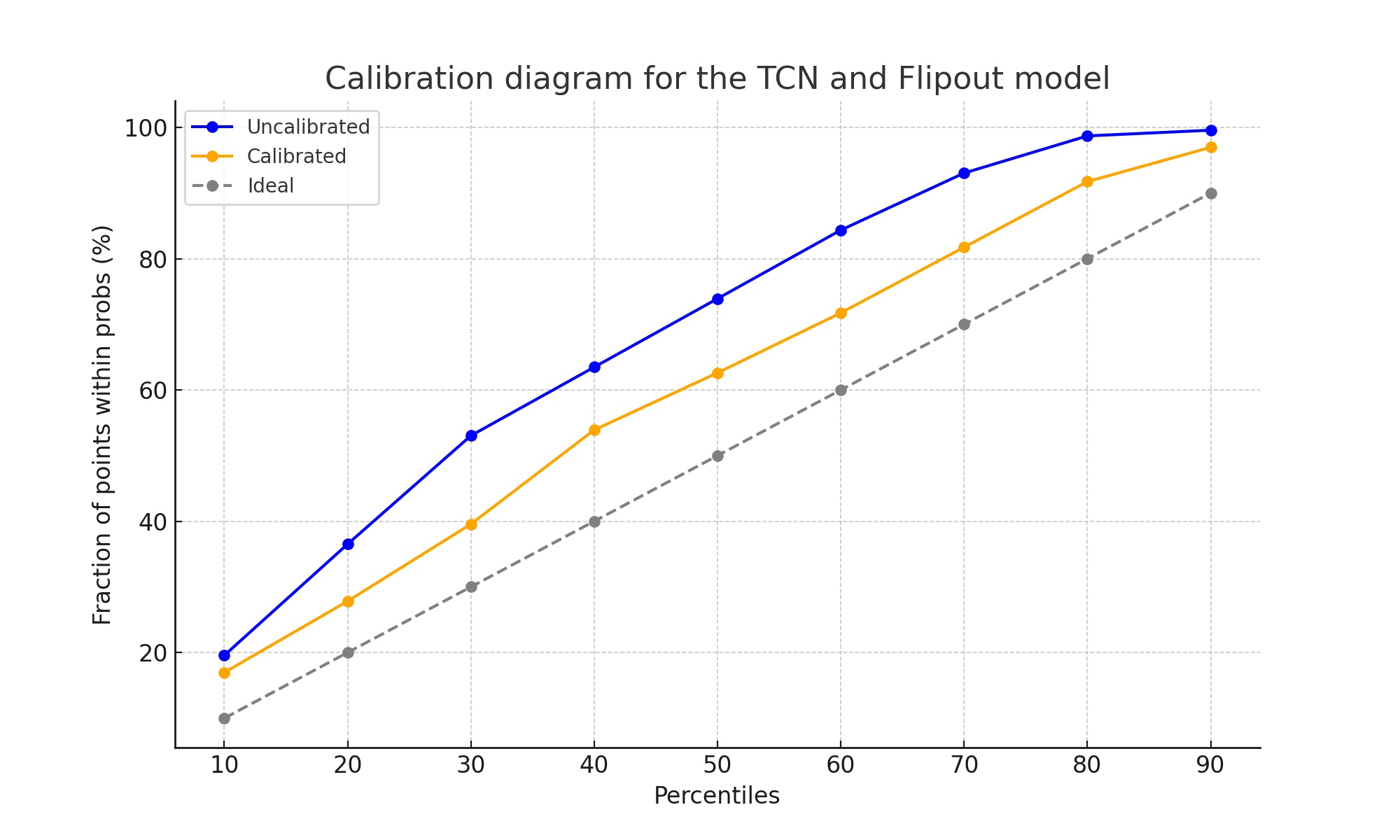}}
\caption{Calibration diagram for the TCN with Flipout model. After minimizing the RMSCE, the scaling factor is equal to 0.7519. The dashed diagonal line represents a perfect calibration.}
\label{fig:tcn_ft_calib}
\end{figure}

\begin{figure}[h!]
\centering
\scalebox{0.7}{\includegraphics[width=\columnwidth]{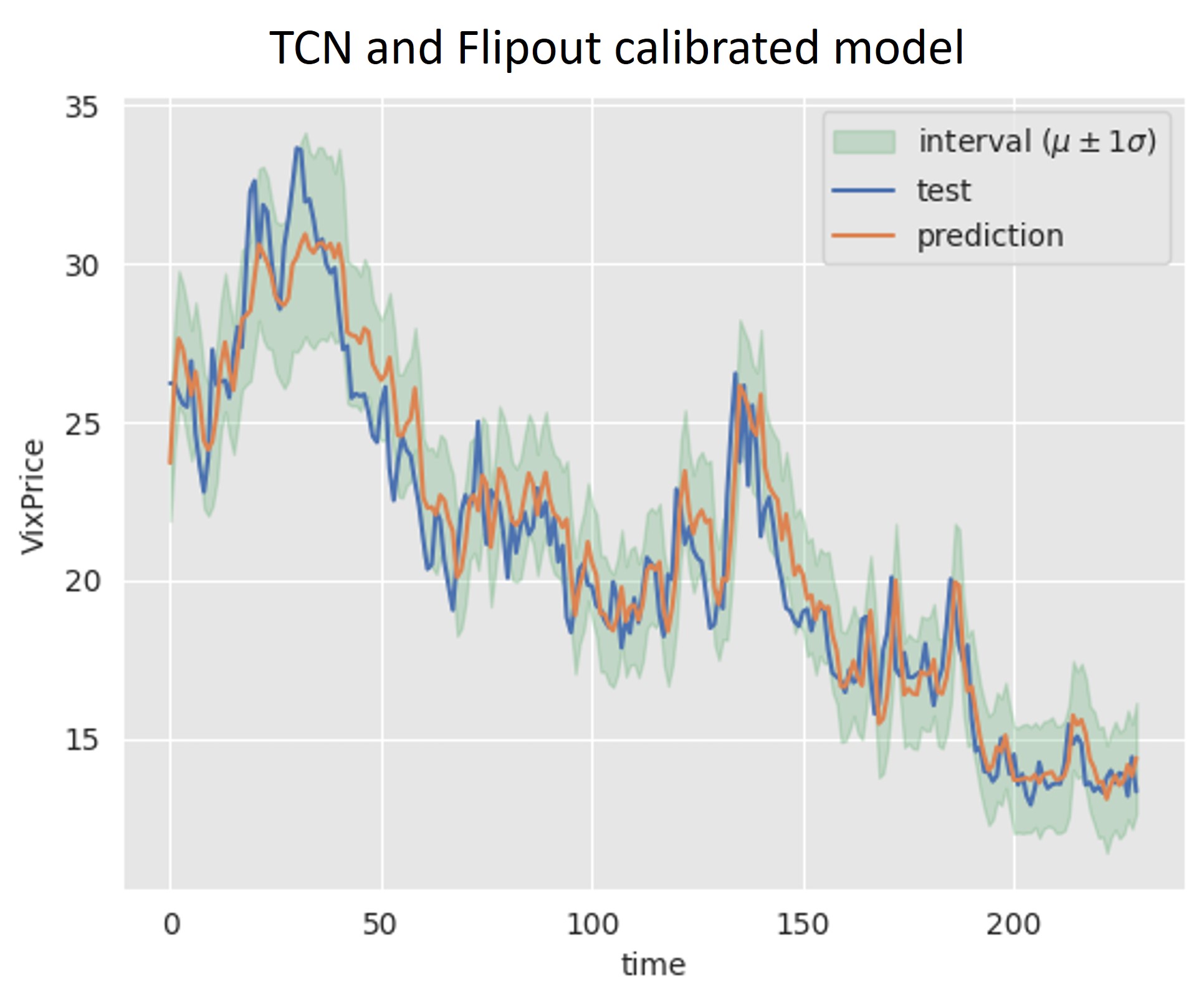}}
\caption{Prediction of the probabilistic TCN and Flipout model for VIX test dataset}\label{fig:tcn_ft_prob}
\end{figure}

\begin{figure}[h!]
\centering
\scalebox{0.8}{\includegraphics[width=\columnwidth]{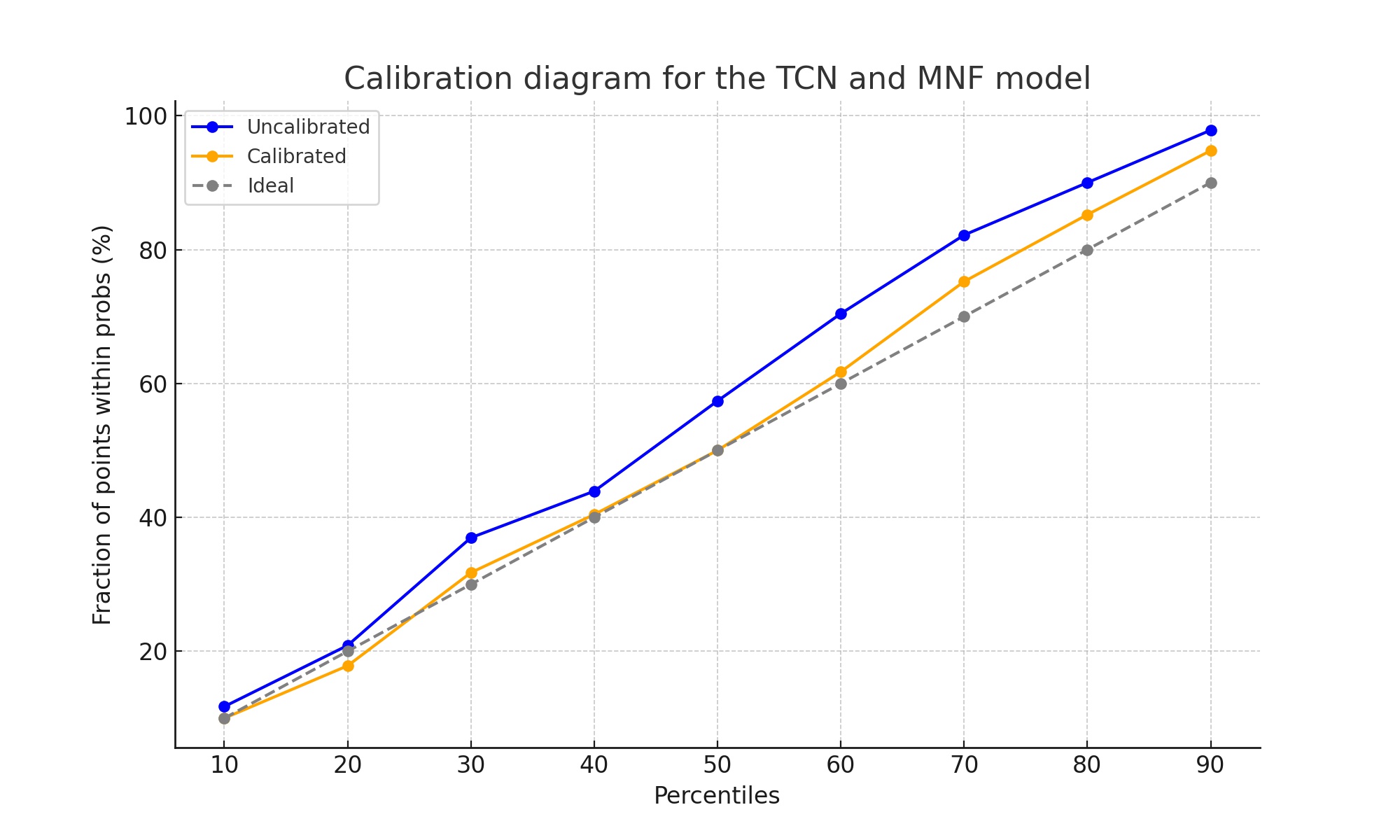}}
\caption{Calibration diagram for the TCN with MNF model. After minimizing the RMSCE, the scaling factor is equal to 0.8825. The dashed diagonal line represents a perfect calibration.}
\label{fig:tcn_mnf_calib}
\end{figure}

\begin{figure}[h!]
\centering
\scalebox{0.7}{\includegraphics[width=\columnwidth]{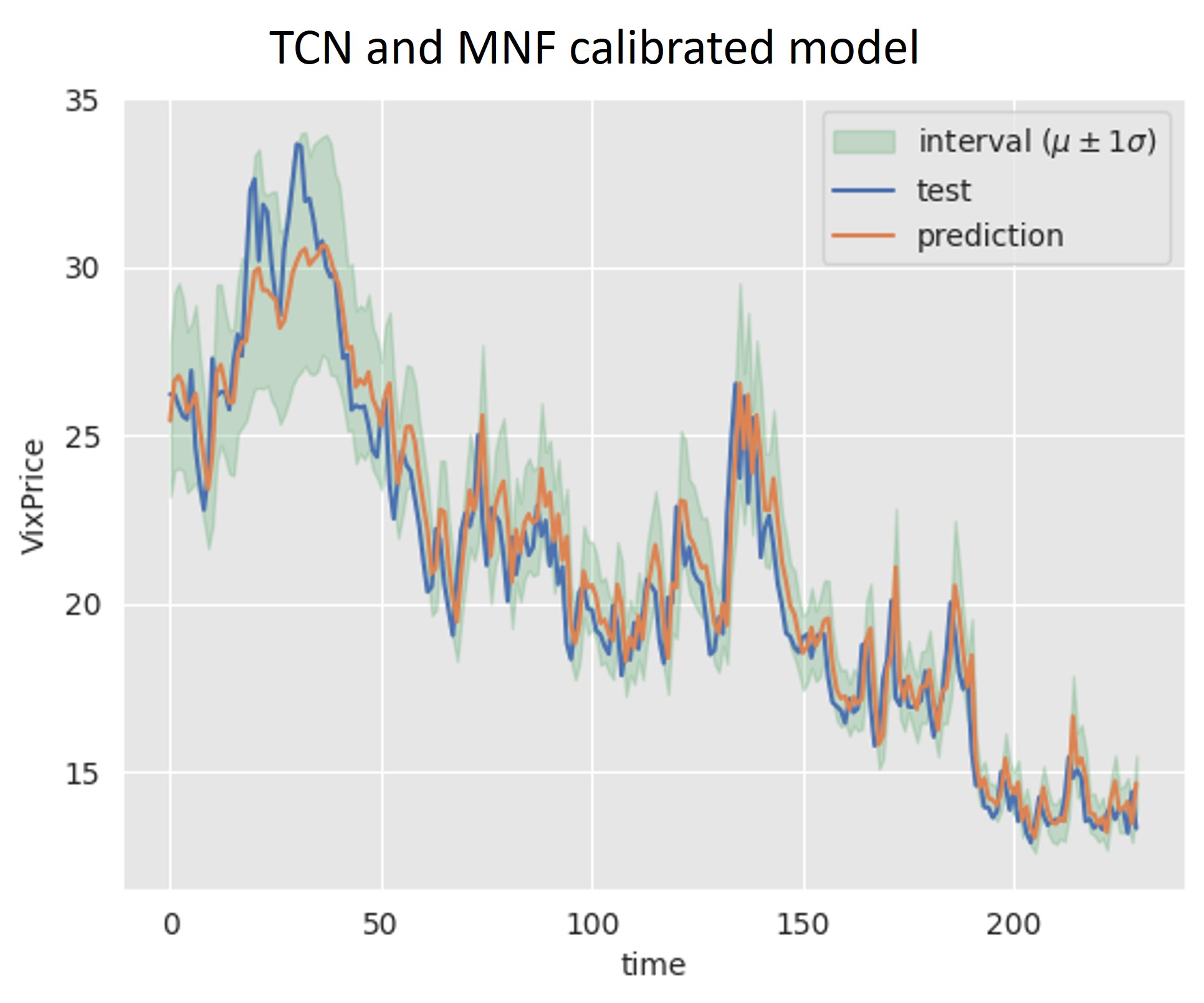}}
\caption{Prediction of the probabilistic TCN and MNF model for VIX test dataset}\label{fig:tcn_mnf_prob}
\end{figure}

\begin{figure}[h!]
\centering
\scalebox{0.8}{\includegraphics[width=\columnwidth]{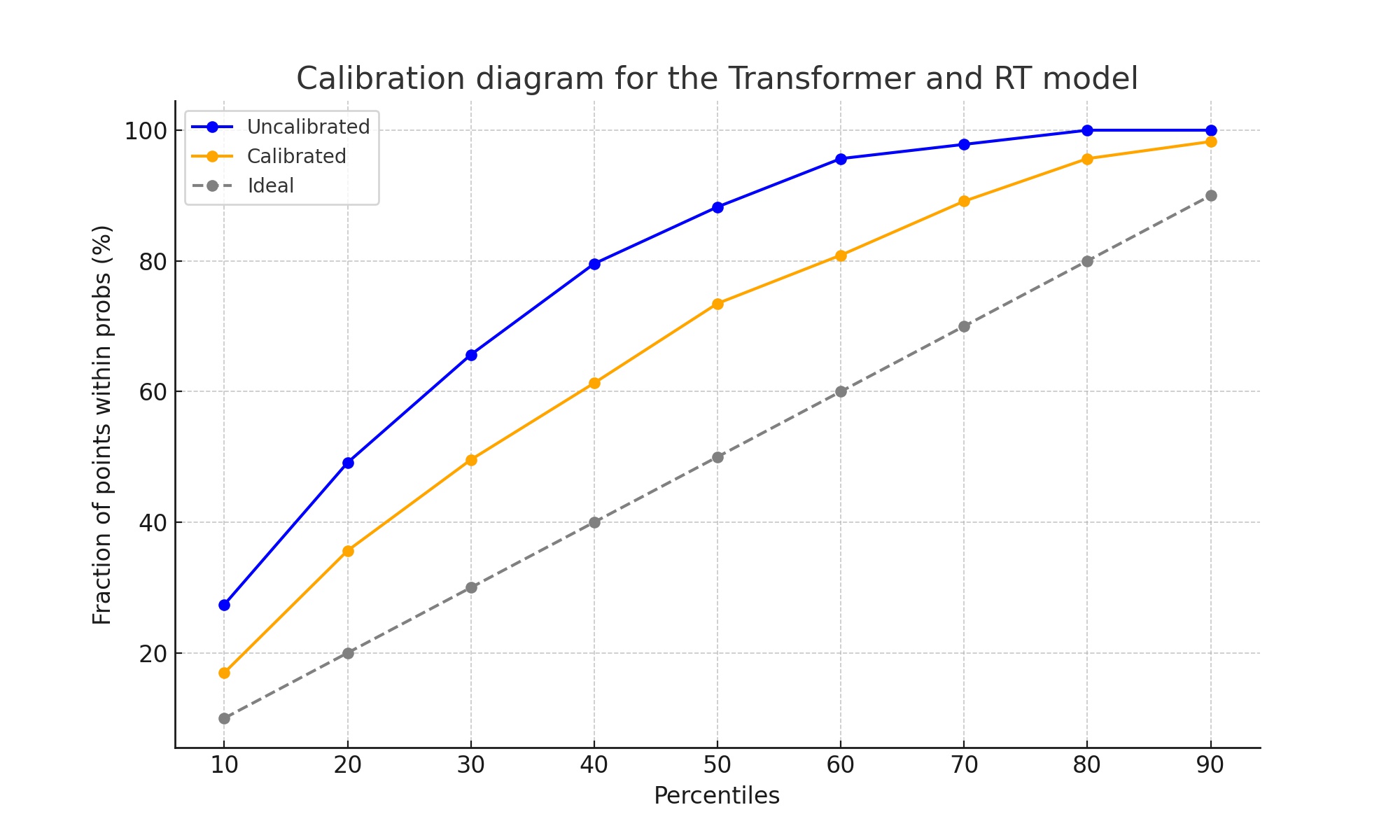}}
\caption{Calibration diagram for the Transformer with RT model. After minimizing the RMSCE, the scaling factor is equal to 0.6699. The dashed diagonal line represents a perfect calibration.}
\label{fig:transf_rt_calib}
\end{figure}

\begin{figure}[h!]
\centering
\scalebox{0.7}{\includegraphics[width=\columnwidth]{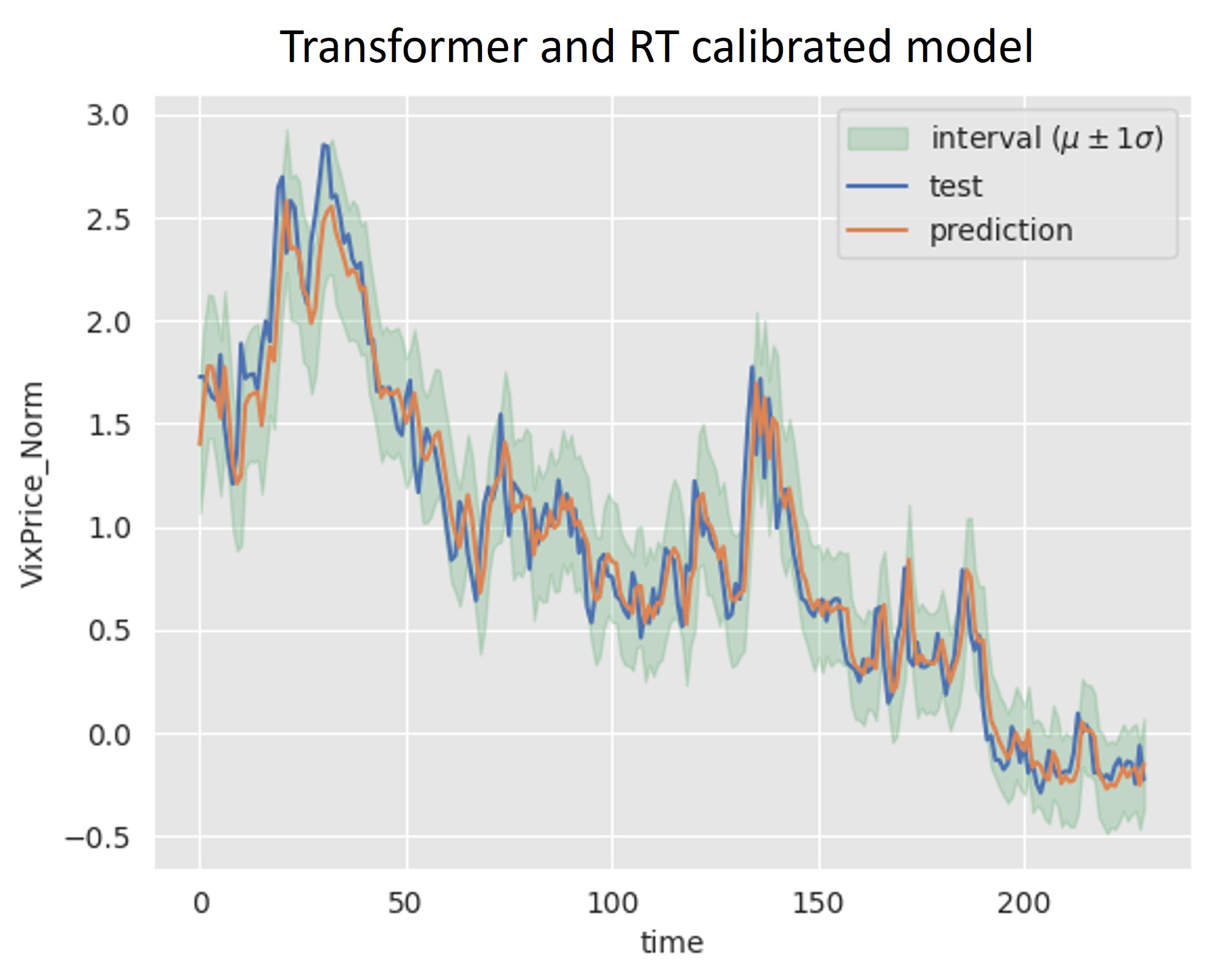}}
\caption{Prediction of the probabilistic Transformer and RT model for VIX test dataset}\label{fig:transf_rt_prob}
\end{figure}

\begin{figure}[h!]
\centering
\scalebox{0.8}{\includegraphics[width=\columnwidth]{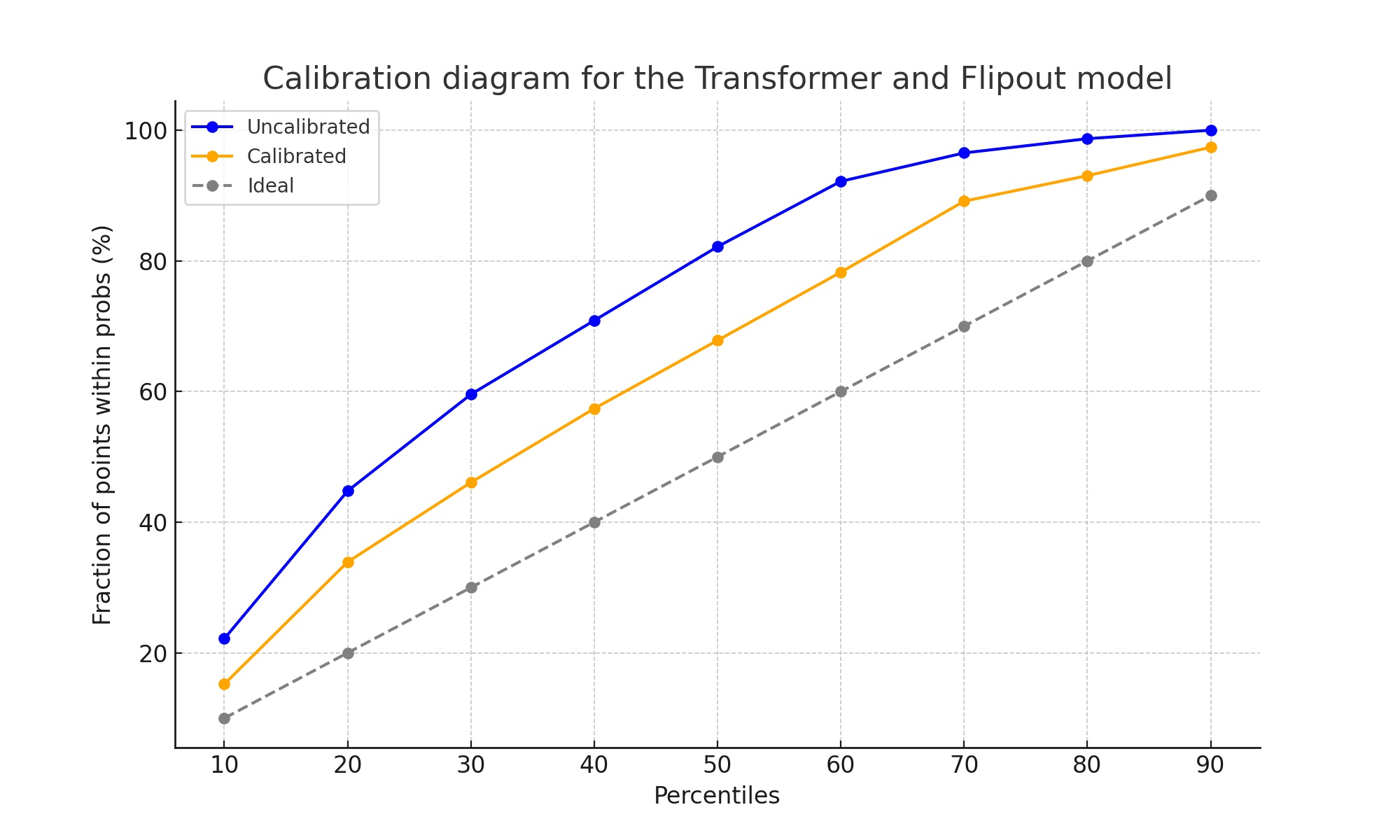}}
\caption{Calibration diagram for the Transformer with Flipout model. After minimizing the RMSCE, the scaling factor is equal to 0.7048. The dashed diagonal line represents a perfect calibration.}
\label{fig:transf_ft_calib}
\end{figure}

\begin{figure}[h!]
\centering
\scalebox{0.7}{\includegraphics[width=\columnwidth]{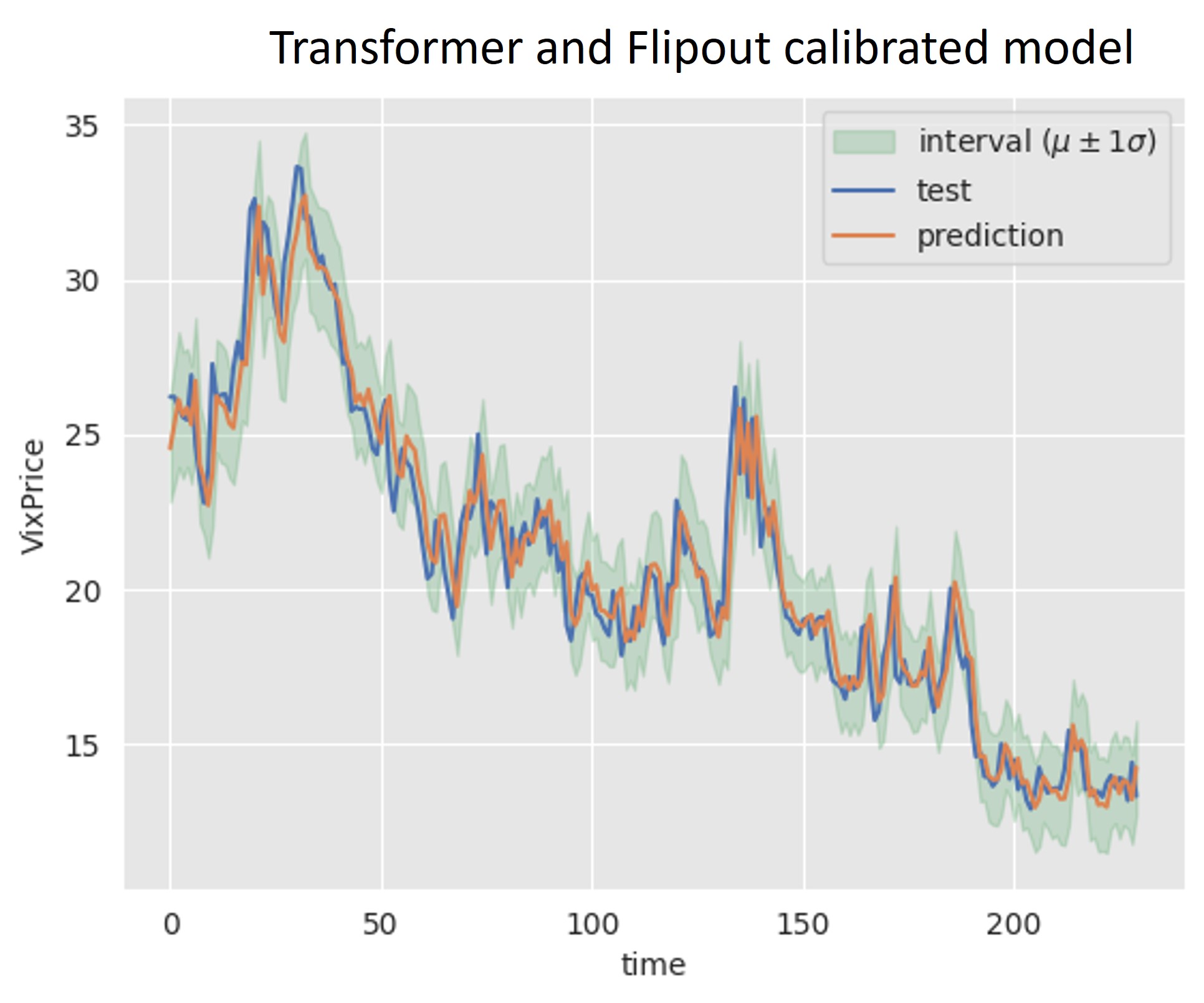}}
\caption{Prediction of the probabilistic Transformer and Flipout model for VIX test dataset}
\label{fig:transf_ft_prob}
\end{figure}

\begin{figure}[h!]
\centering
\scalebox{0.8}{\includegraphics[width=\columnwidth]{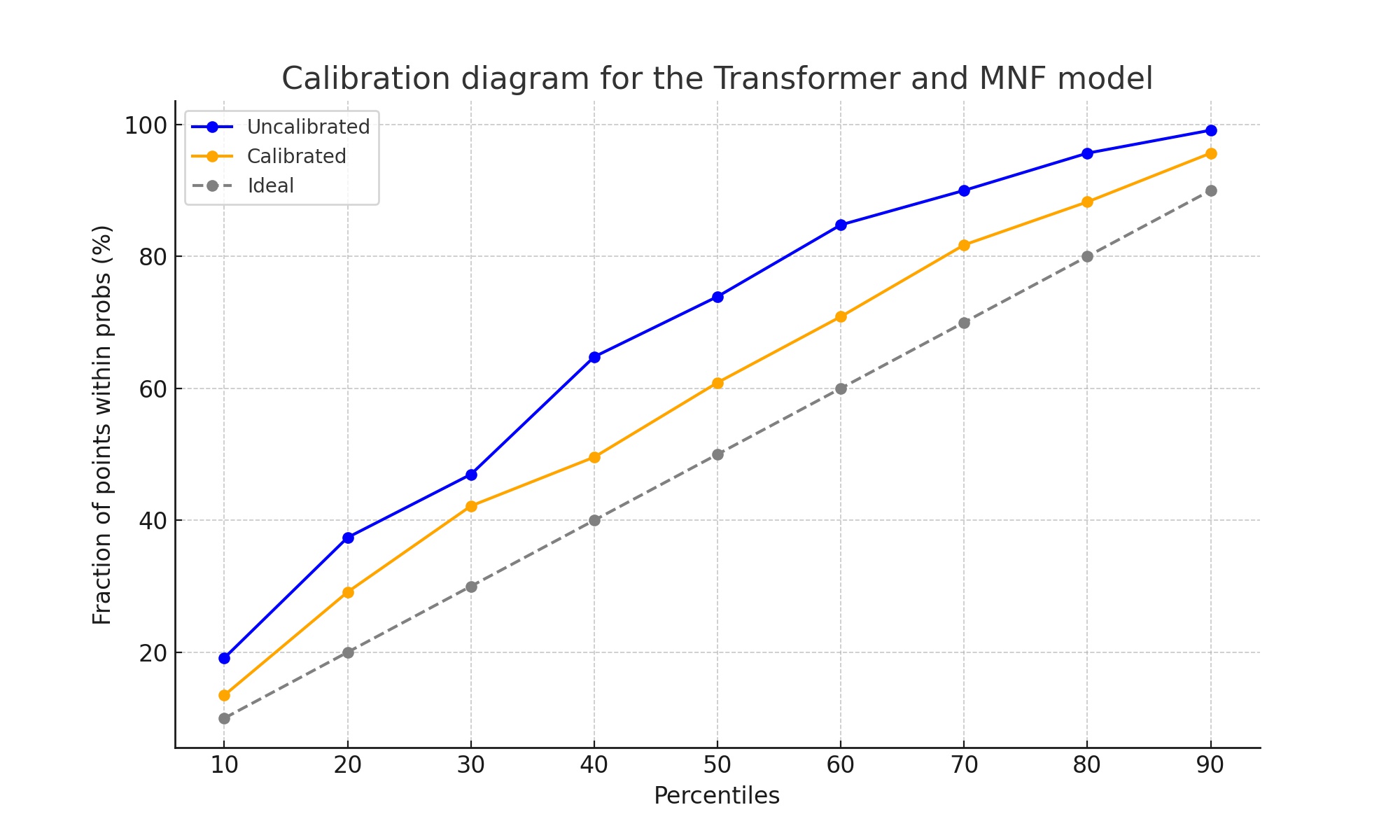}}
\caption{Calibration diagram for the TCN with MNF model. After minimizing the RMSCE, the scaling factor is equal to 0.7641. The dashed diagonal line represents a perfect calibration.}
\label{fig:transf_mnf_calib}
\end{figure}

\begin{figure}[h!]
\centering
\scalebox{0.7}{\includegraphics[width=\columnwidth]{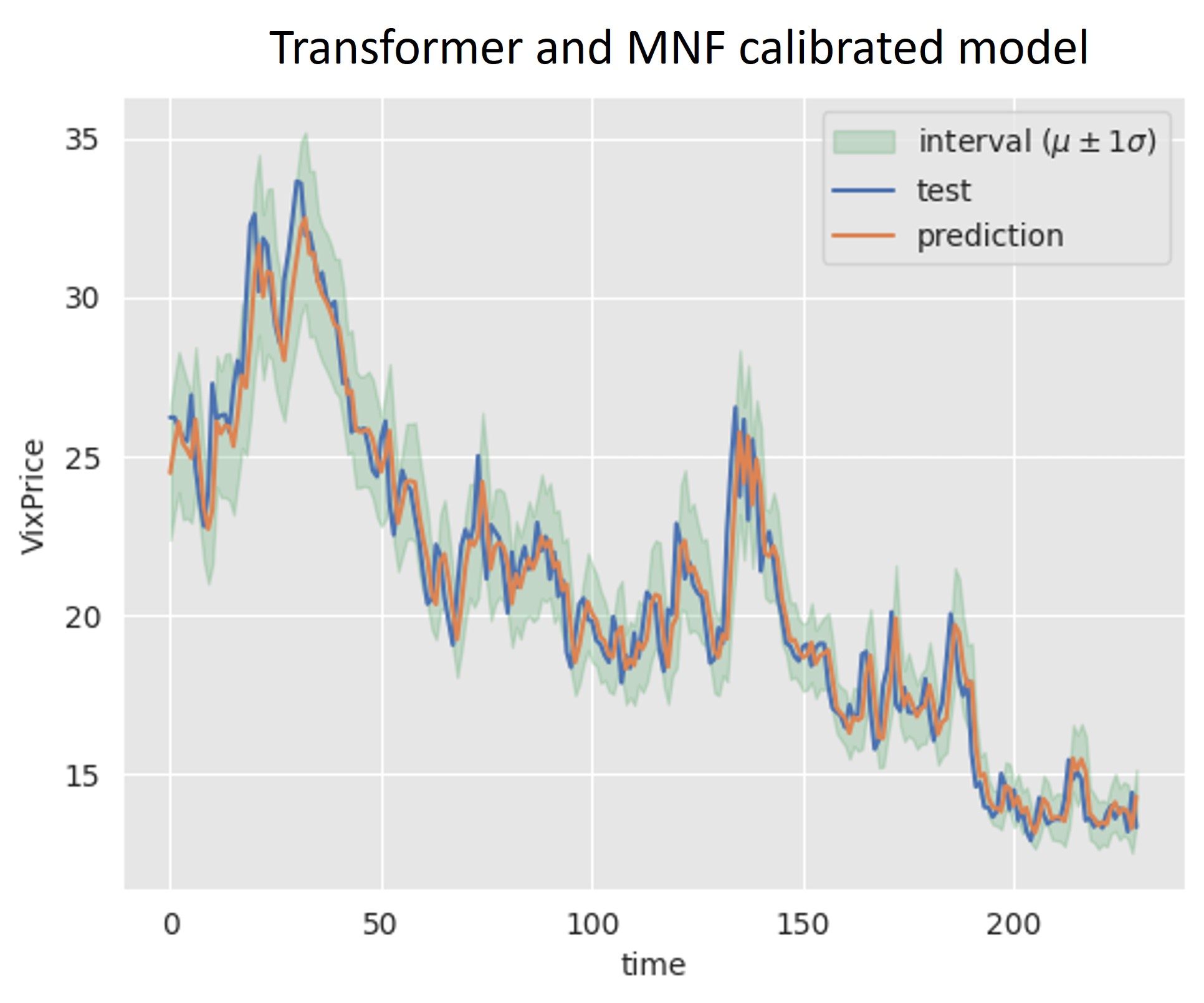}}
\caption{Prediction of the probabilistic Transformer and MNF model for VIX test dataset}\label{fig:transf_mnf_prob}
\end{figure}

It is worth to mention that a scaling factor closer to 1, the better the model, being 1 a perfect calibration.
The initial results of the calibration are shown in Table~\ref{table:result_calib}. 
The MNF (with standard normal prior) presents the higher values of scaling factor and the minimum RMSCE for the three models.
The previous results are confirmed by the calibration diagrams and prediction plots.
Figures~\ref{fig:wn_rt_calib} and \ref{fig:wn_rt_prob} depict the calibration diagram and fit for the WaveNet and RT model.
Whereas, Figures~\ref{fig:wn_ft_calib} and \ref{fig:wn_ft_prob} exhibit the calibration diagram and fit for the WaveNet and Flipout model.
Figures~\ref{fig:wn_mnf_calib} and \ref{fig:wn_mnf_prob} depict the calibration diagram and fit for the WaveNet and MNF model.
On the other hand, Figures~\ref{fig:tcn_rt_calib} and \ref{fig:tcn_rt_prob} show the calibration diagram and fit for the TCN and RT model.
Moreover, Figures~\ref{fig:tcn_ft_calib} and \ref{fig:tcn_ft_prob} present the calibration diagram and fit for the TCN and Flipout model.
Figures~\ref{fig:tcn_mnf_calib} and \ref{fig:tcn_mnf_prob} exhibit the calibration diagram and fit for the TCN and MNF model.
On top of that, Figures~\ref{fig:transf_rt_calib} and \ref{fig:transf_rt_prob} show the calibration diagram and fit for the Transformer and RT model.
Furthermore, Figures~\ref{fig:transf_ft_calib} and \ref{fig:transf_ft_prob} present the calibration diagram and fit for the Transformer and Flipout model.
Finally, Figures~\ref{fig:transf_mnf_calib} and \ref{fig:transf_mnf_prob} depict the calibration diagram and fit for the Transformer and MNF model.

\begin{table}[h!]
\centering
\caption{Results of the initial calibration. A good model has a scaling factor close to 1 and lower values for RMSCE}
\label{table:result_calib}
\begin{tabular}{lcc}
\hline
\textbf{Model} & \textbf{Scaling factor} & \textbf{RMSCE} \\
\hline
\multicolumn{3}{c}{Wavenet}                   \\
RT            & 0.7343                 & 0.0850        \\
Flipout       & 0.7392                 & 0.0916        \\
MNF           & 0.8836                 & 0.0319        \\
\multicolumn{3}{c}{TCN}                      \\
RT            & 0.8589                 & 0.0412        \\
Flipout       & 0.7519                 & 0.0775        \\
MNF           & 0.8825                 & 0.0201        \\
\multicolumn{3}{c}{Transformers}                      \\
RT            & 0.6699                 & 0.1259        \\
Flipout       & 0.7048                 & 0.1048        \\
MNF           & 0.7641                 & 0.0772        \\
\hline
\end{tabular}
\end{table}

\subsection{The Role of Priors}

\begin{figure}[h!]
\centering
\scalebox{1.0}{\includegraphics[width=\columnwidth]{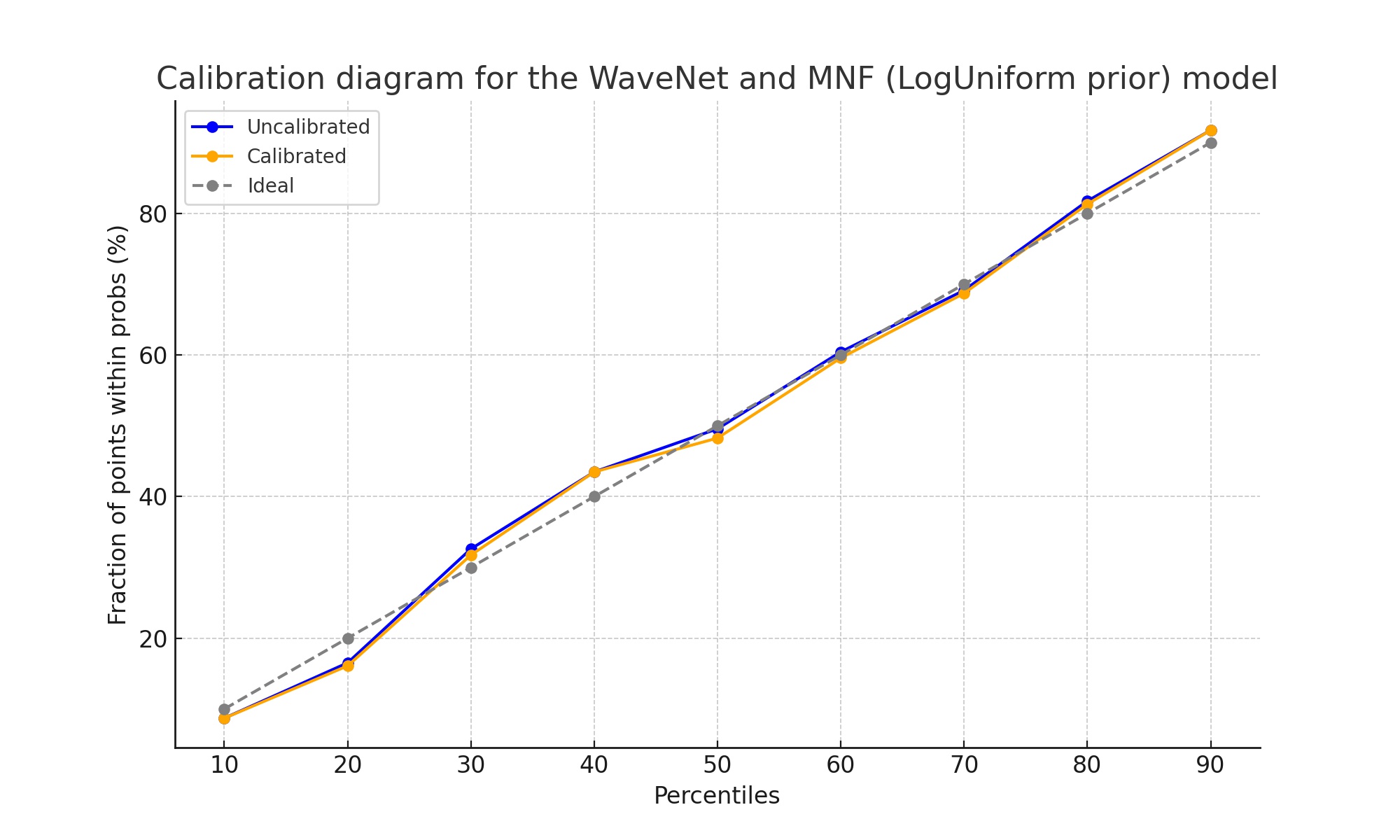}}
\caption{Calibration diagram for the WaveNet with MNF model and LogUniform prior. After minimizing the RMSCE, the scaling factor is equal to 0.9859, meaning a well calibrated network. The dashed diagonal line represents a perfect calibration.}
\label{fig:wn_calib}
\end{figure}

\begin{figure}[h!]
\centering
\scalebox{0.7}{\includegraphics[width=\columnwidth]{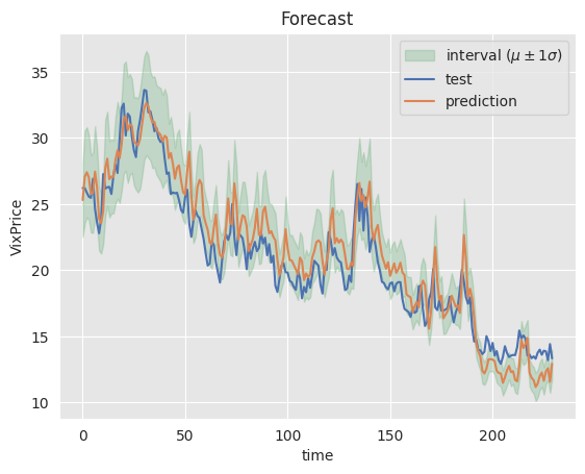}}
\caption{Prediction of the probabilistic WaveNet model for VIX test dataset}\label{fig:wn_prob}
\end{figure}

\begin{figure}[h!]
\centering
\scalebox{1.0}{\includegraphics[width=\columnwidth]{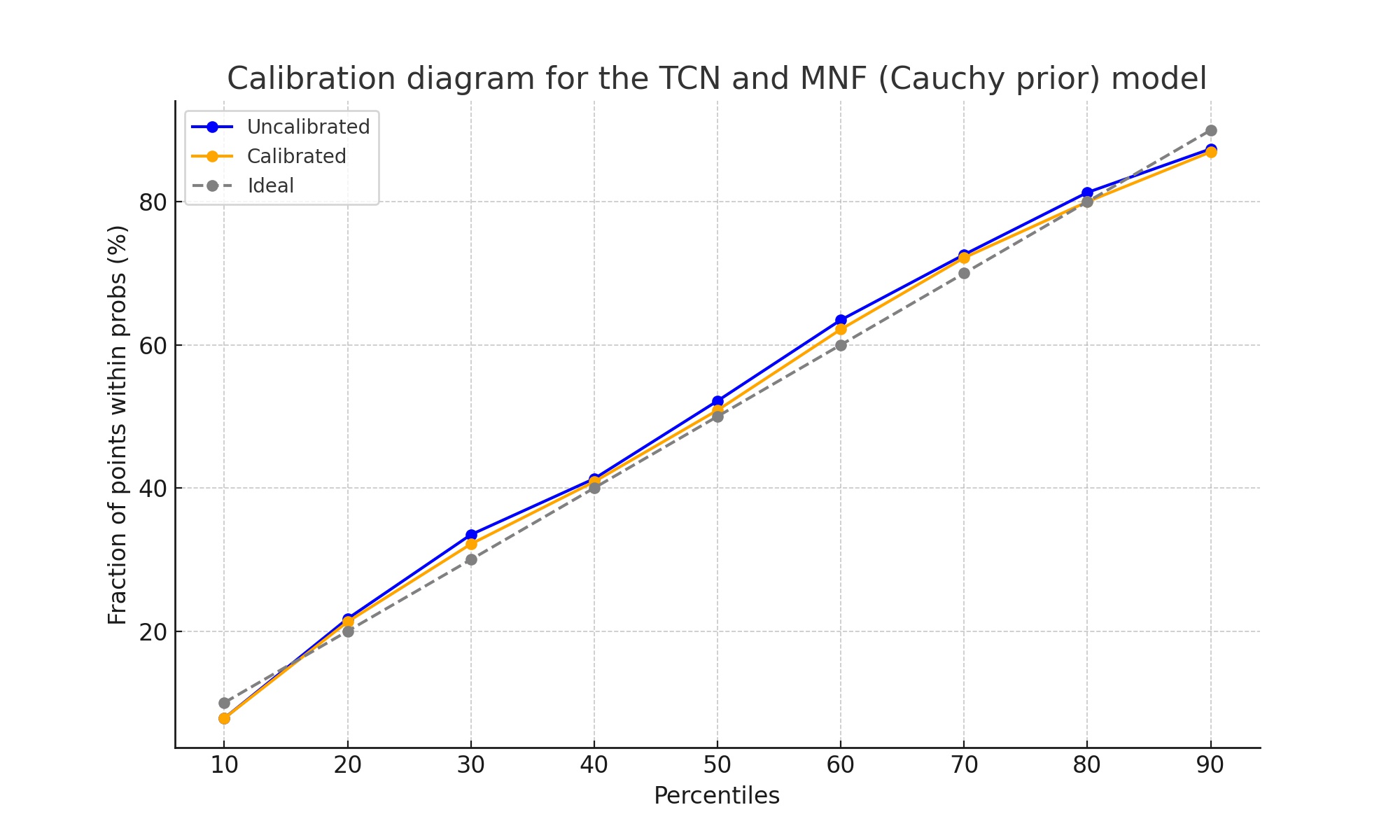}}
\caption{Calibration diagram for the TCN with MNF model and Cauchy prior. After minimizing the RMSCE, the scaling factor is equal to 0.9800, meaning a well calibrated network. The dashed diagonal line represents a perfect calibration.}
\label{fig:tcn_calib}
\end{figure}

\begin{figure}[h!]
\centering
\scalebox{0.7}{\includegraphics[width=\columnwidth]{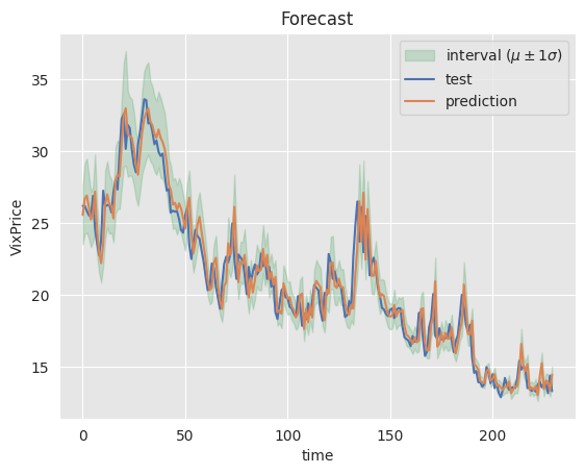}}
\caption{Prediction of the probabilistic TCN with MNF model and Cauchy prior for VIX test dataset. A good point estimate is observed and a higher uncertainty for higher values of VIX, i.e., at the beginning of the graph}\label{fig:tcn_prob}
\end{figure}

The most common distribution for the prior is the normal pdf, but better posterior approximation may be obtained by varying the prior. In our study, we also tested the Cauchy and Log-uniform pdf's (see Table~\ref{table:KL_divergence}).
By changing to these prior distributions in the MNF setup, better results are obtained. For the TCN, the Cauchy distribution prior and two hidden layers with 50 units each, the scaling factor is 0.9800. Whereas for the WaveNet, a scaling factor of 0.9859 is achieved with LogUniform prior and three hidden layers with 50 units each layer. Figure~\ref{fig:wn_calib} shows the calibration diagram for the WaveNet and MNF model (with LogUniform prior) and its prediction after calibration is presented in Figure~\ref{fig:wn_prob}. Whereas, Figure~\ref{fig:tcn_calib} and Figure~\ref{fig:tcn_prob} exhibit the calibration diagram for the TCN and MNF model (with Cauchy prior) and its prediction after the calibration procedure, respectively.

\section{Key Takeways}

All in all, the main results of our work are:

\begin{itemize}
\item It was confirmed that more robust neural networks provide a good forecasting performance for the volatility index VIX in a deterministic and probabilistic setup (as in other many-to-many sequence data), but these networks are miscalibrated \cite{guo17}.
\item MNF with standard normal prior provides better results than RT and Flipout for the calibration procedure in our case study, and 
\item By varying the priors with heavier-tailed distributions in the MNF model, a well calibration is found for the different networks. This is in line with the outstanding works of Fortuin and his team on BNN priors, see for instance \cite{fortuin21} and \cite{fortuin22}.
\end{itemize}

More application works will be needed to compare the performance of uninformative priors (like standard normal)
with heavy-tailed prior distributions and our work shed some lights about the
study of different priors on BNN in the financial time series field.

\begin{table}[h!]
\centering
\caption{KL divergence terms used for the different priors in the MNF model.}
\label{table:KL_divergence}
\begin{tabular}{ll}
\hline
\textbf{Prior} & \textbf{\(-KL\)} \\
\hline
Standard normal & \( \frac{1}{2} [-\log \sigma^2 + \sigma^2 + z^2_{T_f} \mu^2 - 1] \) \\
Log uniform & \( k_1\sigma (k_2 + k_3 \log \tau) - \frac{1}{2} \log(1 + \tau^{-1}) + C \) \\
Standard Cauchy & \( \log \frac{\pi}{2} + \frac{1}{2} [-\log \sigma^2 + \sigma^2 + z^2_{T_f} \mu^2] \) \\
\hline
\end{tabular}
\footnotetext{Source: \href{https://github.com/akashrajkn/waffles-and-posteriors/blob/development/report.pdf}{akashrajkn-priors}}
\end{table}


\begin{figure}[h!]
\centering
\scalebox{1.0}{\includegraphics[width=\columnwidth]{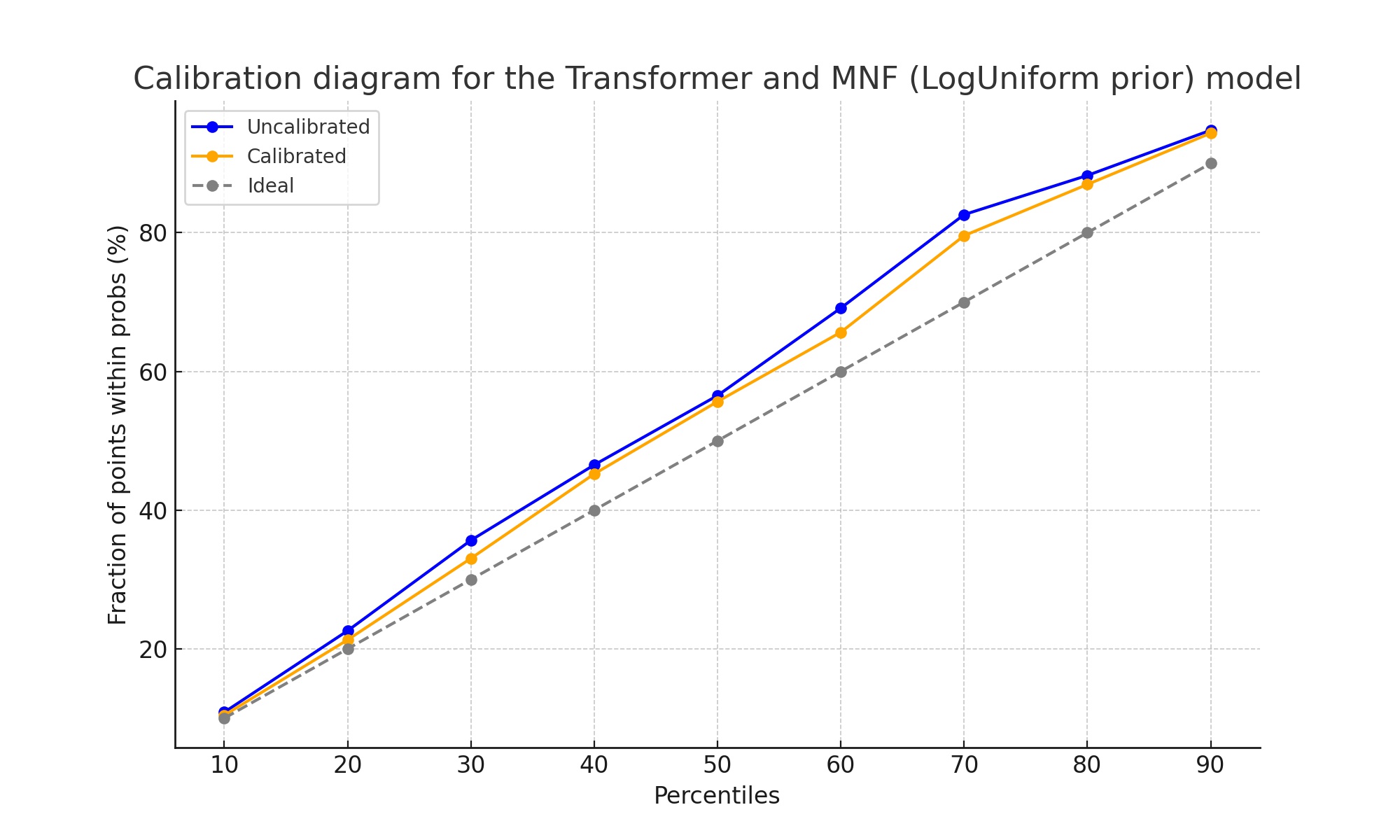}}
\caption{Calibration diagram for the Transformer with MNF model and LogUniform prior. After minimizing the RMSCE, the scaling factor is equal to 0.9418, meaning a well calibrated network. The dashed diagonal line represents a perfect calibration.}
\label{fig:transf_prob}
\end{figure}

\begin{figure}[h!]
\centering
\scalebox{0.7}{\includegraphics[width=\columnwidth]{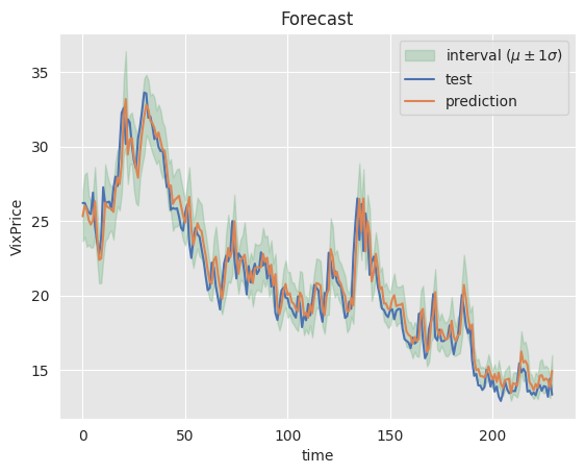}}
\caption{Prediction of the probabilistic Transformer with MNF model and LogUniform prior for VIX test dataset. A good point estimate is observed in general for VIX values}
\label{fig:transf_pred}
\end{figure}


\section{Conclusions and Future Research}\label{sec9}

We implemented Bayesian Neural Networks (BNN) to forecast the volatility index VIX in a probabilistic manner, and thus estimate the weights of two robust neural networks, used in sequence data, like WaveNet, TCN, and Transformer. Three different approaches were employed to this aim, Reparameterization Trick (RT), Flipout, and Multiplicative Normalizing Flows (MNF).  Since modern networks are miscalibrated we employed a simple approach to calibrate the models following the standard deviation scaling method. Our results show that MNF presents the best calibration and overperformance is obtained varying the prior distributions, which is a promising future research in financial time series forecasting with BNN.

Other methodologies related to the analyzed models in our study can be tested such as the
Knowledge-Driven Temporal Convolutional Network (KDTCN) proposed by \cite{deng19} 
who include background knowledge, news and asset price information into deep
prediction models, to mitigate the problem of asset trend forecasting and abrupt changes
explainability. Another model is the Seq-U-Net, where \cite{stoller19} claim is more efficient than other convolutional setups (including TCN and WaveNet).
In the same vein, the Retentive Networks (RetNet), which reduce the inference cost and memory complexity issues of transformer models \cite{sun23}, may be also tested. 
Furthermore, the probabilistic view may be applied to calculate value-at-risk (VaR), which is considered a high quantile of a financial loss distribution,
and contrast results with \cite{mohe20} approach.


\bibliography{sn-bibliography}


\newpage

\begin{appendices}

\section{Additional Graphs for VIX}\label{secA1}

\begin{figure}[h!]
\centering
\scalebox{0.7}{\includegraphics[width=\columnwidth]{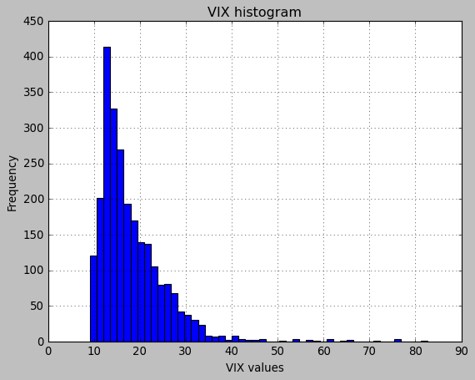}}
\caption{VIX histogram. The analyzed VIX values exhibit a positive skewed distribution with maximum of 82.7 on March 2020 as a consequence of Covid-19 pandemic.}
\label{fig:histogram_vix}
\end{figure}

\begin{figure}[h!]
\centering
\scalebox{0.7}{\includegraphics[width=\columnwidth]{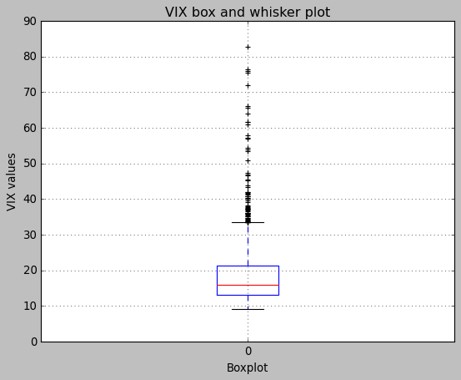}}
\caption{Box and whisker plot. Outliers may be indentified above the VIX value of 40 and the Interquantile Range (IQR) is 8.11.}
\label{fig:box_vix}
\end{figure}

\begin{figure}[h!]
\centering
\scalebox{0.7}{\includegraphics[width=\columnwidth]{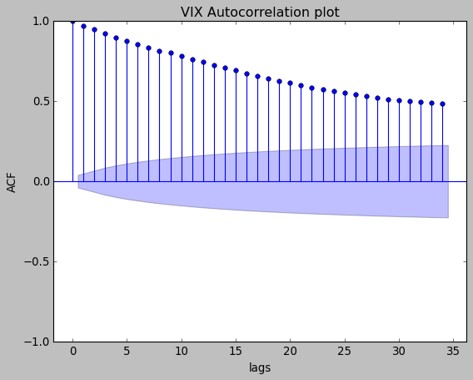}}
\caption{VIX Autocorrelation Function Plot. Long-term dependence behavior can be observed in the VIX values.}
\label{fig:acf}
\end{figure}

\begin{figure}[h!]
\centering
\scalebox{0.7}{\includegraphics[width=\columnwidth]{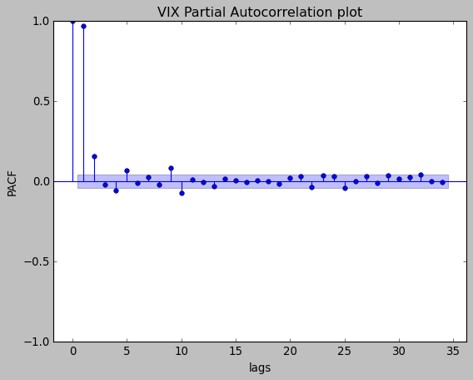}}
\caption{VIX Partial Autocorrelation Function Plot. PACF measures the remaining correlation after eliminating the correlation effect in between. Together with the ACF plot, an AR(2) may be identified for the VIX time series.}
\label{fig:pacf}
\end{figure}


\end{appendices}

\end{document}